%% file: bare_jrnl.tex
\documentclass[journal]{IEEEtran}
\usepackage{tikz}
\usepackage{amsmath}
\usepackage{comment}
\usepackage{color}
\usepackage{booktabs}
\usepackage{xspace}
\usepackage{amssymb}
\usepackage{amsfonts}           
\usepackage{mathrsfs} 
\usepackage{subfigure}
\usepackage{multirow}
\usepackage{booktabs}
\usepackage{graphicx}
\usepackage{enumitem}
\newcommand{\mymodel}{Deep-Masking Generative Network\xspace}
\newcommand{\peicomment}[1]{\textcolor[rgb]{1,0,0} {#1}}
\newcommand{\maRevise}[1]{\textcolor[rgb]{0,0,0} {#1}}
\newcommand{\miRevise}[1]{\textcolor[rgb]{0,0,0} {#1}}
\ifCLASSINFOpdf
\else
\fi
\begin{document}
%

\title{\mymodel: A Unified Framework for Background Restoration from Superimposed Images}

%
%
%

\author{Xin Feng,
        Wenjie Pei$^*$,
        Zihui Jia,
        Fanglin chen,~\IEEEmembership{Member,~IEEE},
        David Zhang,~\IEEEmembership{Life~Fellow,~IEEE} 
        and~Guangming Lu$^*$,~\IEEEmembership{Member,~IEEE}
\thanks{This work was supported by the NSFC fund (62006060 and U2013210), the Guangdong Basic and Applied Basic Research Foundation (2019Bl515120055, 2021A1515012528), the Shenzhen Fundamental Research Fund (JCYJ20180306172023949), the Open Project Fund (AC01202005018) from Shenzhen Institute of Artificial Intelligence and Robotics for Society, and the Medical Biometrics Perception and Analysis Engineering Laboratory, Shenzhen, China.}
\thanks{$^*$Wenjie Pei and Guangming Lu are corresponding authors.}
\thanks{Xin Feng, Wenjie Pei, Fanglin Chen and Guangming Lu are with the Department of
Computer Science, Harbin Institute of Technology at Shenzhen, Shenzhen
518057, China (e-mail: fengx\text{\_hit}@outlook.com; wenjiecoder@outlook.com; chenfanglin@hit.edu.cn; luguangm@hit.edu.cn).}
\thanks{Zihui Jia is with Tencent (e-mail: xibeijia@tencent.com).}
\thanks{David Zhang is with the School of Science and Engineering, The Chinese University of Hong Kong at Shenzhen, Shenzhen 518172, China (e-mail: davidzhang@cuhk.edu.cn)}
}

\maketitle

\begin{abstract}
\input{abstract}
\end{abstract}

\begin{IEEEkeywords}
Image background restoration, reflection removal, rain streak removal, image dehazing, residual deep-masking cell.
\end{IEEEkeywords}

%
\IEEEpeerreviewmaketitle

%
%
%
%



\section{Introduction}
\input{introduction.tex}

\vspace{-2pt}
\section{Related Work}
\input{related_work.tex}

\vspace{-2pt}
\section{\mymodel}
\input{method.tex}

\vspace{-2pt}
\section{Experiments}
\input{experiments.tex}

\vspace{-2pt}
\section{Conclusion}
\input{conclusion}

\ifCLASSOPTIONcaptionsoff
  \newpage
\fi

%
 \bibliographystyle{IEEEtran}
\bibliography{mybibfile}
\vspace{-20pt}
\begin{IEEEbiography}[{\includegraphics[width=1in,height=1.25in,clip,keepaspectratio]{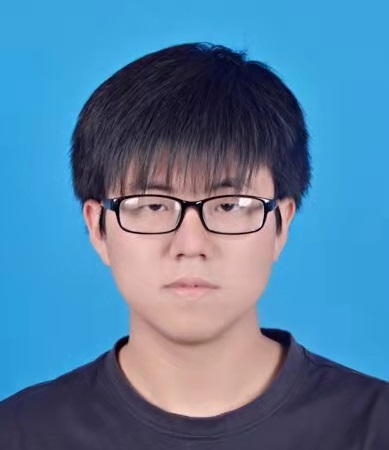}}]{Xin Feng} received the B.S. degree from Zhengzhou University, China in 2017. He is currently pursuing the Ph.D. degree with the Harbin Institute of Technology, Shenzhen, China. His current research interests include low-level vision, deep learning, generative model, and relevant applications.
\end{IEEEbiography}

\begin{IEEEbiography}[{\includegraphics[width=1in,height=1.25in,clip,keepaspectratio]{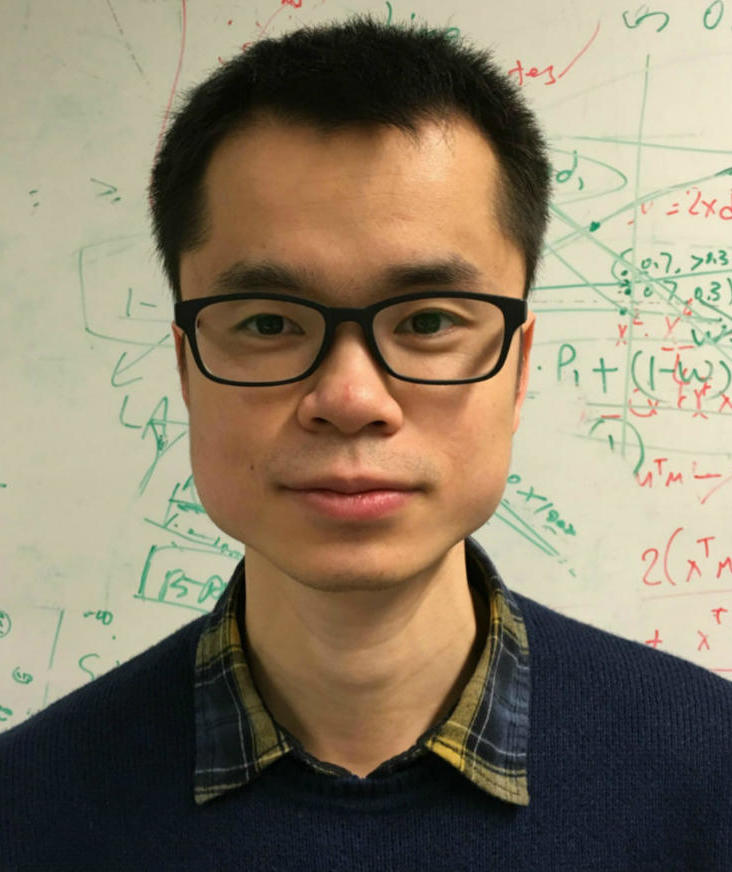}}]{Wenjie Pei}
is currently an Assistant Professor with the Harbin Institute of Technology, Shenzhen, China. He received the Ph.D. degree from the Delft University of Technology, the Netherlands in 2018, working with Dr. Laurens van der Maaten and Dr. David Tax. Before joining Harbin Institute of Technology, he was a Senior Researcher on Computer Vision at Tencent Youtu X-Lab. In 2016, he was a visiting scholar with the Carnegie Mellon University. His research interests lie in Computer Vision and Pattern Recognition including sequence modeling, deep learning, video/image captioning, etc.
\end{IEEEbiography}

\begin{IEEEbiography}[{\includegraphics[width=1in,height=1.25in,clip,keepaspectratio]{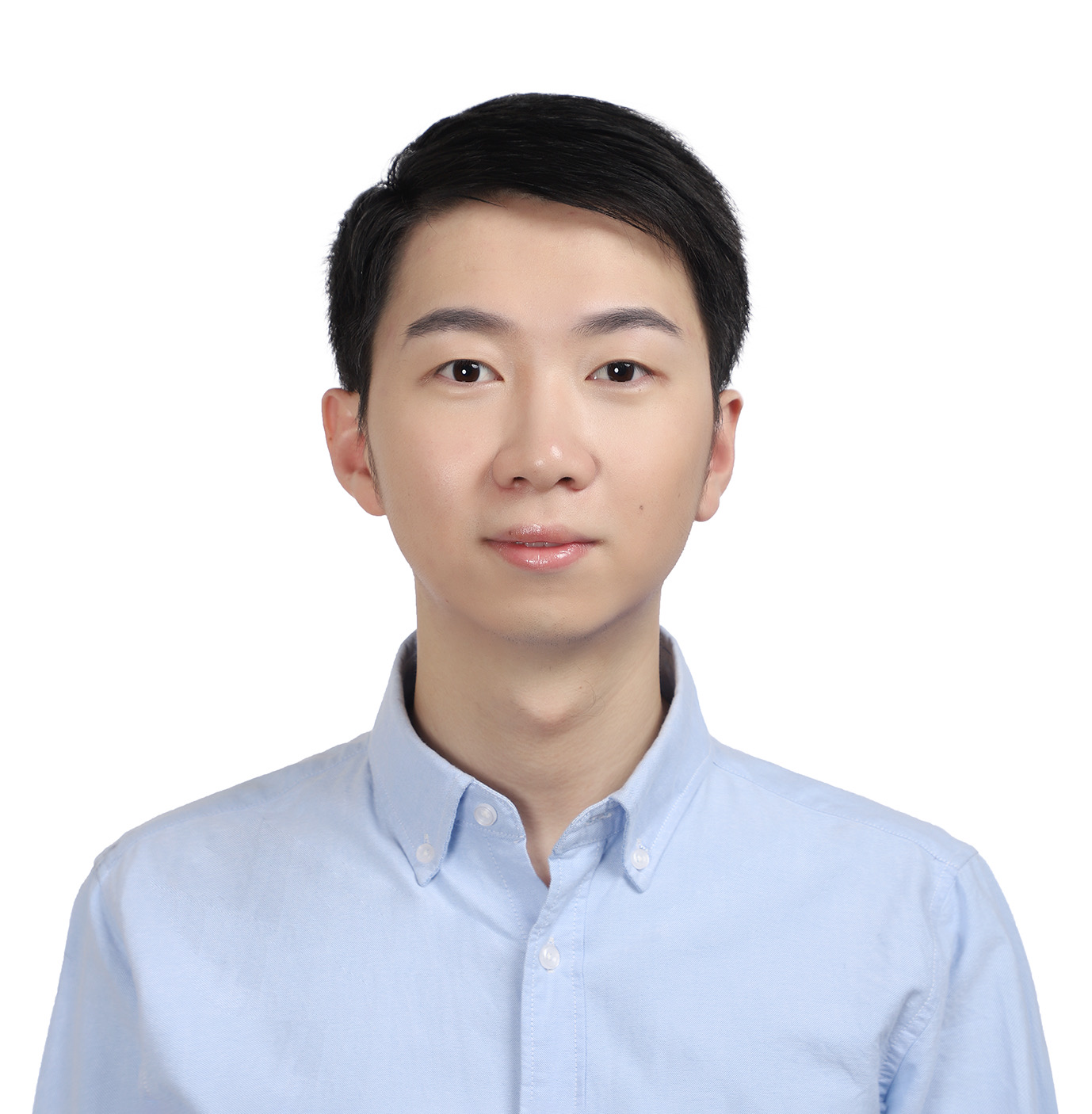}}]{Zihui Jia}
received the B.S. degree in information and software engineering from the University of Electronic Science and Technology of China in 2017, and the M.S. degree in computer science from the Peking University Shenzhen Graduate School in 2020. He is now working as a researcher in Tencent Youtu Lab and his research interests include image processing, object detection, and semantic segmentation.
\end{IEEEbiography}

\begin{IEEEbiography}[{\includegraphics[width=1in,height=1.25in,clip,keepaspectratio]{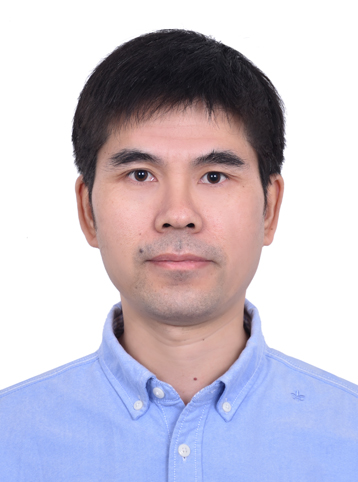}}]{Fanglin Chen}
	received the B.S. degree in control science and engineering and the B.S. degree in applied mathematics from Xi’an Jiaotong University, Xi’an, China, in 2006. In 2011, he received the Ph.D. degree in control science and engineering from Tsinghua University, Beijing, China. He is currently a Professor with the Department of Computer Science and Technology, Harbin Institute of Technology, Shenzhen. His current research areas include  computer vision, pattern recognition, and image processing.  He received the Academic Newcomer Award (Ministry of Education, China) in 2010, and the First Group of Robot Vision Challenge (ImageCLEF) in 2014.
\end{IEEEbiography}	

\begin{IEEEbiography}[{\includegraphics[width=1in,height=1.25in,clip,keepaspectratio]{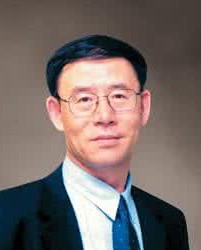}}]{David Zhang}(Life Fellow, IEEE) graduated in Computer Science from Peking University. He received his MSc in 1982 and his PhD in 1985 in both Computer Science from the Harbin Institute of Technology (HIT), respectively. From 1986 to 1988 he was a Postdoctoral Fellow at Tsinghua University and then an Associate Professor at the Academia Sinica, Beijing. In 1994, he received his second PhD in Electrical and Computer Engineering from the University of Waterloo, Ontario, Canada. Currently, he is a Professor at Chinese University of Hong Kong (Shenzhen). He also serves as Visiting Chair Professor in Tsinghua University and HIT, and Adjunct Professor in Shanghai Jiao Tong University, Peking University, National University of Defense Technology and the University of Waterloo. He is the Founder and Editor-in-Chief, International Journal of Image and Graphics (IJIG); Book Editor, Springer International Series on Biometrics (KISB); Organizer, the first International Conference on Biometrics Authentication (ICBA); Associate Editor of more than ten international journals including IEEE Transactions and so on. He has published over 20 monographs, 400 international journal papers and 40 patents from USA/Japan/HK/China. Professor Zhang is a Croucher Senior Research Fellow, Distinguished Speaker of the IEEE Computer Society, and a Fellow of both IEEE and IAPR.
\end{IEEEbiography}

\begin{IEEEbiography}[{\includegraphics[width=1in,height=1.25in,clip,keepaspectratio]{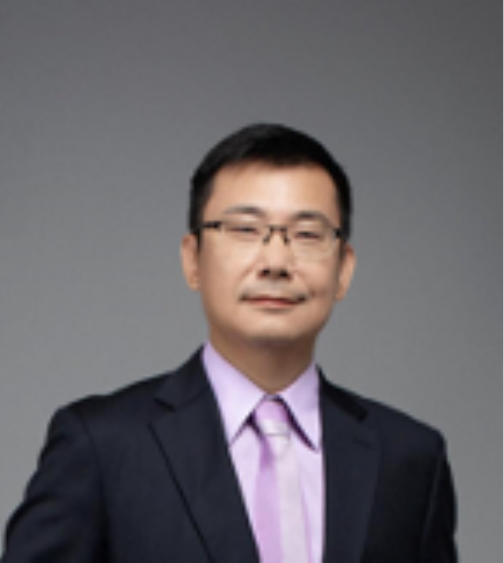}}]{Guangming Lu}
received the undergraduate degree in electrical engineering, the master degree in control theory and control engineering, and the Ph.D. degree in computer science from the Harbin Institute of Technology (HIT), Harbin, China, in 1998, 2000, and 2005, respectively. From 2005 to 2007, he was a Postdoctoral Fellow at Tsinghua University. Now, He is a Professor at Harbin Institute of Technology, Shenzhen, China. He has published over 100 technical papers at some international journals and conferences, including IEEE TIP, TNNLS, TCYB, TCSVT, NeurIPS, CVPR, AAAI, IJCAI, etc. His research interests include computer vision, pattern recognition, and machine learning.
\end{IEEEbiography}

%

\end{document}

%% file: abstract.tex
Restoring the clean background from the superimposed images containing a noisy layer is the common crux of a classical category of tasks on image restoration such as image reflection removal, image deraining and image dehazing. These tasks are typically formulated and tackled individually due to diverse and complicated appearance patterns of noise layers within the image.
In this work we present the Deep-Masking Generative Network (\emph{DMGN}), which is a unified framework for background restoration from the superimposed images and is able to cope with different types of noise. 

Our proposed \emph{DMGN} follows a coarse-to-fine generative process: a coarse background image and a noise image are first generated in parallel, then the noise image is further leveraged to refine the background image to achieve a higher-quality background image. In particular, we design the novel Residual Deep-Masking Cell as the core operating unit for our \emph{DMGN} to enhance the effective information and suppress the negative information during image generation via learning a gating mask to control the information flow. By iteratively employing this Residual Deep-Masking Cell, our proposed \emph{DMGN} is able to generate both high-quality background image and noisy image progressively. Furthermore, we propose a two-pronged strategy to effectively leverage the generated 
noise image as contrasting cues to facilitate the refinement of the background image. Extensive experiments across three typical tasks for image background restoration, including image reflection removal, image rain steak removal and image dehazing, show that our \emph{DMGN} consistently outperforms state-of-the-art methods specifically designed for each single task.

%% file: introduction.tex
\IEEEPARstart{B}{ackground} restoration from superimposed images is an important yet challenging research problem involving many tasks of image restoration, such as image reflection removal, image deraining and image dehazing. The key characteristic of this type of tasks is that the input polluted image can be viewed as a superimposed image composed of a background layer and a noise layer. The crux of these tasks stems from the complexities of diverse noise patterns across different tasks that are exceedingly difficult to recognize. Thus, existing methods typically formulate these tasks as \miRevise{individual} research problems and address them \miRevise{separately}. Consequently, a well-designed model for one of these tasks is hardly applied to another one \miRevise{directly}. In this work we aim to propose a unified solution for background restoration from superimposed images, \miRevise{which is able to cope with various tasks involving different types of noise patterns like reflection, rain, or haze.}

\begin{figure*}[t]
\centering
\includegraphics[width=0.9\linewidth]{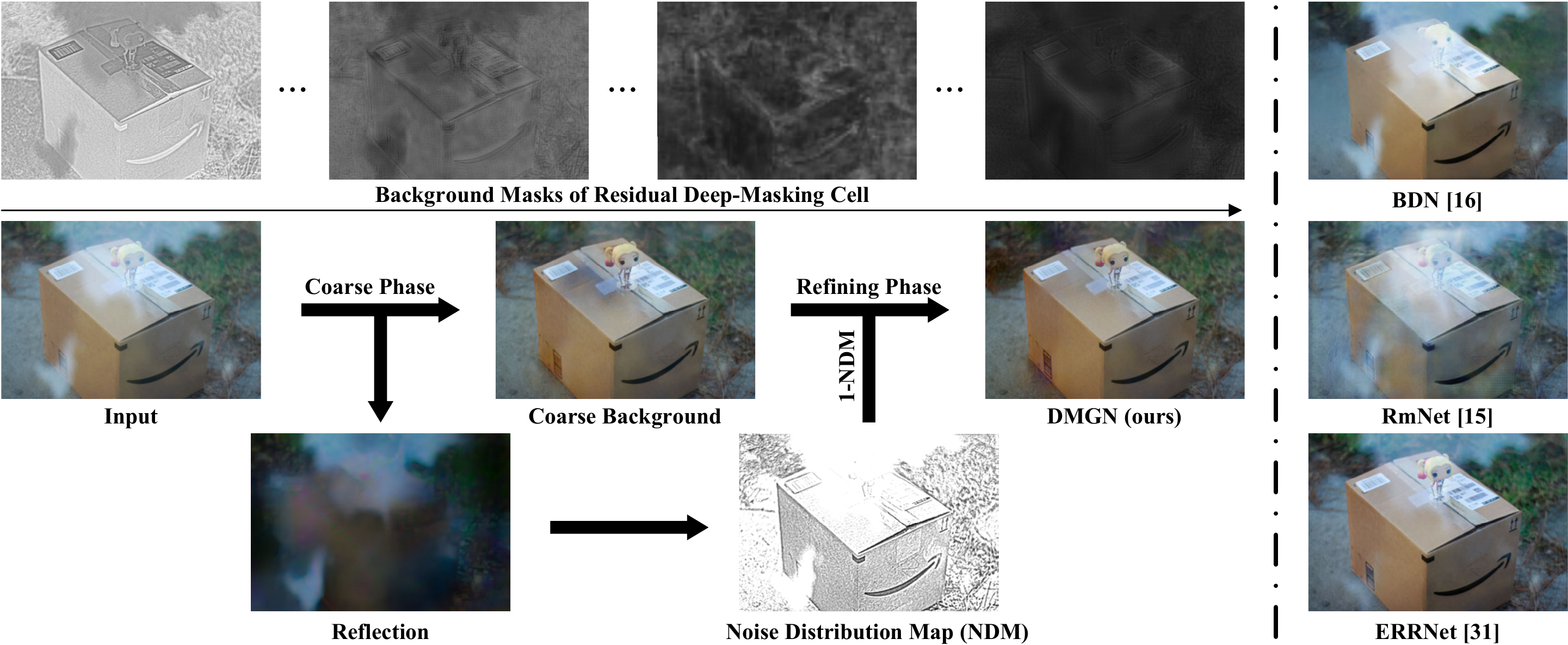}
\vspace{-4pt}
\caption{Given an input image with reflection, our \emph{DMGN} first generates a coarse background image and a noise (reflection in this task) image in the coarse phase by iteratively employing the proposed Residual Deep-Masking Cell, then the generated reflection image is further utilized to refine the background image. Here Residual Deep-Masking Cell learns a gating mask to control information propagation in each iteration during image generation (darker value indicate less information flow): enhance the effective information and suppress the negative information. As the proceeding of the background generation, the restoration of background is gradually finished and less information is required. Hence, more and more pixel values in the mask tend to converge to 0. Consequently, our \emph{DMGN} is able to generate a cleaner background image than other methods.}
\vspace{-4pt}
\label{fig:intro}
\end{figure*}

A prominent way of restoring background from superimposed images adopted by traditional methods~\cite{chen2013generalized,li2016rain,zhu2017joint,levin2007user,levin2004separating,wan2016depth} is to leverage prior knowledge about noise to model the noise layer and then \miRevise{restore} the background image by contrasting with the \miRevise{obtained} noise layer. Whilst \miRevise{such} modeling paradigm has been validated its effectiveness in tasks with easy-to-model noise patterns like rain streak removal~\cite{chen2013generalized,li2016rain,zhu2017joint}, it fails to deal with tasks with complicated and diverse noise patterns such as reflection or haze. 

With great success in various tasks of computer vision by the powerful feature representation, deep convolutional networks (CNNs) were first employed for background restoration from noise images by simply stacking plain convolutional layers~\cite{zhang2018single,fu2017clearing,cai2016dehazenet}. It is expected to model the mapping function from the noise image to background image in a brute-force way by CNNs, which turns out to be hardly achieved due to the complicated noise patterns. 
\miRevise{This limitation is alleviated by recursively employing a same designed CNN-based module and improve the quality of the synthesized background image recurrently~\cite{li2018recurrent,ren2019progressive,li2019single,fu2019lightweight}.} While such method indeed boosts the performance of image background restoration to some degree, it is at the expense of more computation and consuming time. 

Another research line of improving the performance of image background restoration is to synthesize the background image and the noise image in parallel and then leverage the obtained noise image to facilitate the background synthesis~\cite{ma2019learning,wen2019single,yang2018seeing,fu2017removing,li2017single,yang2017deep,zhang2018density}, which follows two typical ways. A straightforward way is to concatenate the synthesized noise image and the input image together to generate a higher-quality background image~\cite{yang2018seeing,fu2017removing,li2017single,yang2017deep,zhang2018density}. However, it is difficult for a parameterized (deep learning) model to automatically learn the implicit cues from the synthesized noise image by simply concatenating it to the input image. Another typical way of leveraging the predicted noise image is to reconstruct the input image from both the generated background image and the noise image, and then perform supervision on the whole model by a reconstruction loss~\cite{ma2019learning,wen2019single}. Nevertheless, such method brings little improvement probably because of the uncertainty and error introduced during the reconstruction process. 

In this work we propose the Deep-Masking Generative Network (\emph{DMGN}), which is a unified framework for background restoration from superimposed images and is able to handle different types of image noise such as reflection, rain streak or haze. It follows the coarse-to-fine generative process. To be specific, it first synthesizes a coarse background image and a noise image in the coarse phase \miRevise{in parallel}, then the generated noise image is further leveraged to refine the coarse background image and finally obtain a higher-quality background image in the refining phase. Compared to existing methods for image background restoration for various related tasks, our \emph{DMGN} benefits from two advantages:
\begin{itemize}
    \item A \textbf{Residual Deep-Masking Cell (RDMC)} is designed specifically as a core operating unit for \emph{DMGN} to enhance the effective information and suppress the negative information during image synthesis. \miRevise{It learns a gating mask based on attention mechanism to control the information propagation for each pixel of feature maps}. Thus, the model restores the background image and the noise image by learning information masks to summarize the noise patterns dynamically by itself based on the training data rather than the prior knowledge concerning the noise patterns, which enables the model to have more generalization and adaptability to diverse noise patterns across different tasks. By iteratively stacking this Residual Deep-Masking Cell rather than the plain convolutional layers, \emph{DMGN} is able to generate both high-quality background image and noise image in a progressively-refining manner.
\item We propose a \textbf{two-pronged strategy to effectively leverage the generated noise image} as explicit contrasting cues to facilitate the refinement of the coarse background image. In particular, we utilize the generated noise image to regularize both the source information and the masking mechanism of the Residual Deep-Masking Cell during the refinement.
\end{itemize}

Consider the example of image reflection removal \miRevise{presented} in Figure~\ref{fig:intro}, our \emph{DMGN} first generates a coarse background image and reflection (noise) image employing the Residual Deep-Masking Cell from the input image. Then the \miRevise{generated} reflection image is further leveraged to refine the background image by learning a reflection distribution map as a contrasting cues. Consequently, our \emph{DMGN} is able to restore much cleaner and higher-quality background image than other methods. 
We conduct extensive experiments on three different tasks including image reflection removal, image rain streak removal and image dehazing, to evaluate our \emph{DMGN}, demonstrating that \emph{DMGN} consistently outperforms state-of-the-art methods specifically designed for each single task on both \emph{PSNR} and \emph{SSIM} metrics~\cite{wang2004image}.

%% file: related_work.tex
While a full review covering all various tasks of image background restoration is infeasible due to the space limitation, in this section we first summarize the generic methods commonly adopted across different tasks of image background restoration, then we review the recent related work of three typical tasks on which we conduct experiments to evaluate our model: (1) image reflection removal, (2) image rain streak removal and (3) image dehazing. \miRevise{Finally we review the existing methods which perform image background restoration in a unified framework and compare these methods with our proposed method.} 

\smallskip\noindent\textbf{Generic methods across different tasks of image background restoration.} Three types of generic methods for image background restoration can be concluded. \miRevise{The first type of generic methods is the prior-based method, which restores background image by leveraging the physical properties in degradation} such as edge blurring~\cite{arvanitopoulos2017single,li2014single,6823128}, streak directionality\cite{jiang2017novel,jiang2018fastderain,garg2004detection}, or atmosphere scattering modeling\cite{he2010single,cai2016dehazenet}. Another type of generic methods perform background restoration in a recursive manner~\cite{li2018recurrent,ren2019progressive,li2019single,fu2019lightweight}. \miRevise{It recursively employs a designed restoration module and performs supervision in each iteration, aiming to progressively refine background image}. The third type of generic methods first synthesize the noise image and then refine the background image based on the obtained noise image~\cite{ma2019learning,wen2019single,yang2018seeing,fu2017removing,li2017single,yang2017deep,zhang2018density}.

\smallskip\noindent\textbf{Image reflection removal.} Reflection removal is a typical background restoration task. Its goal is \miRevise{to perform content decomposition on reflection-contaminated image to obtain the background image and the reflection image}. Fan \emph{et al.}~\cite{fan2017generic} first \miRevise{introduces} a two-stage convolution neural network for reflection removal, which \miRevise{first} predicts the background edge and then utilizes it to achieve the reflection-free background. Afterwards, an architecture with two cooperative sub-networks~\cite{wan2018crrn} is proposed, which concurrently predicts the background edge and background layer. Then, Yang \emph{et al.}~\cite{yang2018seeing} use a multi-stage architecture to learn implicit cues between different layers without specific guide. Zhang \emph{et al.}~\cite{zhang2018single} propose to employ perceptual loss and dilated convolution for better details restoration. Later, Wei \emph{et al.}~\cite{wei2019single} propose an alignment-invariant loss to train misaligned image pairs, which reduces the dependence on real-world images. Besides, some studies on image reflection removal~\cite{wen2019single,ma2019learning} focus on synthesizing realistic images with reflection for data augmentation to alleviate the deficiency of training data.

\smallskip\noindent\textbf{Image rain streak removal.} Fu \emph{et al.}~\cite{fu2017removing} propose the Deep Detail Network (DDN) which focuses on preserving details. Yang \emph{et al.}~\cite{yang2017deep} design a multi-stream network \miRevise{referred to as} JORDER to perform detection and removal of rain streak jointly. Residual-guided feature fusion network(RGN)~\cite{yang2017deep} designs a lightweight cascaded network to remove rain streak progressively. A density-aware multi-stream dense CNN is proposed in~\cite{zhang2018density}, which is combined with generative adversarial network~\cite{goodfellow2014generative} to jointly estimate rain density and removal rain. Besides, PReNet~\cite{ren2019progressive} provides a simple yet effective baseline, which synthesizes the background progressively by a specifically designed recurrent structure. A pivotal state-of-the-art method for rain streak removal is RESCAN~\cite{li2018recurrent}, which follows the classical paradigm of rain streak removal: \miRevise{it first models the rain streak layer and then obtain the background image from the contrast between the rain streak layer the rain image.}

\smallskip\noindent\textbf{Image dehazing.} He~\emph{et al.}~\cite{he2010single} first propose to leverage the dark channel prior(DCP), \miRevise{which is based on the pixel statistic of color channels}, to achieve the background image. However, this method might degrade in the cases of inconspicuous objects. \miRevise{To address this limitation}, Cai \emph{et al.}~\cite{cai2016dehazenet} design an end-to-end CNN model termed as `DehazeNet' to predict medium transmission map to obtain clean background image by atmosphere scattering model. AOD-Net~\cite{li2017aod} reformulates the atmosphere scattering modeling with CNN model to directly obtain dehazing image. \miRevise{A potential limitation of these methods is that} they rely on the atmosphere scattering model \miRevise{heavily}. \miRevise{To eliminate such dependence}, Ren \emph{et al.}~\cite{ren2018gated} propose a gated fusion network(GFN), which estimates three haze residue maps to generate final haze-free image. Chen \emph{et al.}~\cite{chen2019gated} propose a gated context aggregation network(GCANet), which introduces the smoothed dilated operation to effectively solve common grid artifacts.

\smallskip\noindent\maRevise{\textbf{Unified framework for image background restoration.} Two recent work, which has similar motivation as our model, has made attempts to solve different tasks of image background restoration in a unified framework.}
\maRevise{Zou \emph{et al.}~\cite{Zou_2020_CVPR} proposes a unified framework based on adversarial supervision for separating superimposed images. It focuses on designing multi-level discriminators to maximally separate between the background layer and noise layer in a superimposed image. Experiments across different tasks of background restoration in Section~\ref{sec:reflection_exp} show prominent advantages of our model over Zou \emph{et al.}~\cite{Zou_2020_CVPR}.
On the other hand Li \emph{et al.}~\cite{li2020all} designs multiple task-specific encoders, each of which is responsible for handling one particular type of background restoration task (about bad weather). While such mechanism performs well for targeted tasks, the distinct difference from our method is that such method is still designed in a task-oriented manner and it has to design a corresponding encoder for each of targeted tasks. Thus the model has to be updated once there is new targeted task. By contrast, our method performs background restoration without any specific design to each targeted task. Thus the whole model can be directly applied to any task of image background restoration in a task-agnostic manner.
}

%% file: method.tex
\begin{figure*}[!ht]
\centering
\includegraphics[width=0.98\linewidth]{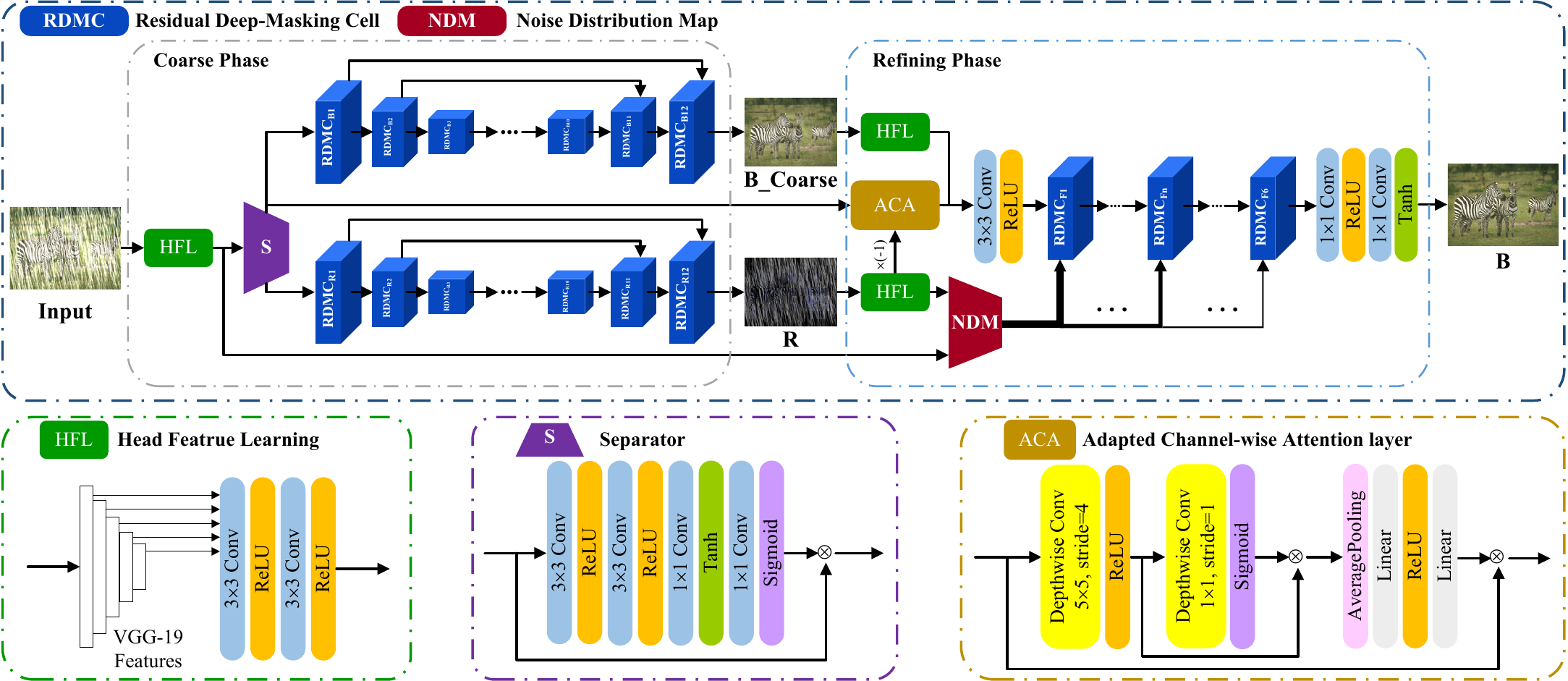}
\vspace{-4pt}
\caption{Architecture of the proposed Deep-Masking Generative Network (\emph{DMGN}). It restores the background image following a coarse-to-fine framework: a coarse background image and a noise image (rain streak image in this example of rain streak removal) are first generated in the coarse phase, then the noise (rain streak) image is further leveraged to refine the coarse background image and achieve a higher-quality background image in the refining phase (presented in Section~\ref{sec:refine} and Figure~\ref{fig:RDMC2}). The Residual Deep-Masking Cell (presented in Section~\ref{sec:RDMC} and Figure~\ref{fig:RDMC}) is specifically designed as the core operating unit for image generation in both the coarse phase and the refining phase.}
\vspace{-4pt}
\label{fig:model}
\end{figure*}

Given an superimposed image containing noise like reflection, rain streak or other forms of noise, we aim to restore it to a noise-free background image. To this end, we propose the Deep-Masking Generative Network (\emph{DMGN}), \miRevise{which is a unified framework for background restoration and is able to deal with various types of noise.} It performs restoration of the background image following a coarse-to-fine framework. A coarse background image and a noise image are first generated in parallel from the input image, then the coarse background image is further refined by contrasting with the generated noise image to obtain a higher-quality background image.
Considering the complexity of various noise patterns across different tasks, our \emph{DMGN} is constructed as a deep-masking architecture which iteratively cascades the core operating unit: Residual Deep-Masking Cell. It is specifically designed to enhance the effective information and suppress the useless information during image generation via maintaining a gating mask. By deeply stacking this Residual Deep-Masking Cell, the proposed \emph{DMGN} is able to generate the background image and the noise image progressively. 

The overall architecture is illustrated in Figure~\ref{fig:model}. We will first introduce the coarse-to-fine generative framework adopted in \emph{DMGN}, then we will elaborate on the Residual Deep-Masking Cell and describe how the coarsely generated background image is refined by contrasting with the generated noise image. Finally, we will show that our \emph{DMGN} can be trained in an end-to-end manner by joint supervision. 
\subsection{Coarse-to-fine Generative Framework}
Our \emph{DMGN} restores the noise-free background image from the input image following a coarse-to-fine framework. As shown in Figure~\ref{fig:model}, it is composed of two generative phases: coarse phase and refining phase. In the coarse phase, the content of the input image is \miRevise{decomposed} into the background constituent and the noise constituent in the deep embedding space. Then a coarse background image and a noise image are synthesized from the corresponding constituent in parallel. In the refining phase, the generated noise image is leveraged as a contrasting cue to further refine the coarse background image. Consequently, a high-quality background image can be restored by the proposed \emph{DMGN}.

Formally, given an image with noise $I$, \maRevise{\emph{DMGN} first projects the input image $I$ into a deep embedding feature space by a feature-learning head, which adopts Hypercolumn technique~\cite{hariharan2015hypercolumns} to aggregate multilevel features using a pre-trained VGG~\cite{simonyan2014very} and a learnable two-layer convolutional module.} Subsequently a separator $\mathcal{S}$ is employed to separate the content of the input image into background constituent and noise constituent, which are then fed into two parallel branches to synthesize a coarse background image and a noise image respectively. Hence, the coarse phase of \emph{DMGN} is formulated as:
\begin{equation}
    \begin{split}
         &\mathbf{F} = \mathcal{F}(I), \\
         &\mathbf{F}_b, \mathbf{F}_r = \mathcal{S}(\mathbf{F}), \\
         &B_{\text{coarse}} = \mathcal{G}_b (\mathbf{F}_b), \\
         &R = \mathcal{G}_r (\mathbf{F}_r),
    \end{split}
\label{eqn:coarse}
\end{equation}
where $\mathbf{F}$ is the obtained feature maps from the feature-learning head $\mathcal{F}$. $\mathbf{F}_b$ and $\mathbf{F}_r$ are the separated background constituent and noise constituent respectively. $B_{\text{coarse}}$ and $R$ are generated coarse background image and the noise image by the background generator $\mathcal{G}_b$ and the noise generator $\mathcal{G}_r$. 

The separator $\mathcal{S}$ is modeled by the spatial attention mechanism to estimate the distribution of background constituent in feature maps $\mathbf{F}$:
\begin{equation}
    \begin{split}
         &\mathbf{A} = \sigma (f_{att}(\mathbf{F})), \\
         &\mathbf{F}_b = \mathbf{F} \odot \mathbf{A}, \\
         &\mathbf{F}_r = \mathbf{F} - \mathbf{F}_b.
    \end{split}
\end{equation}
Herein, $f_{att}$ denotes the nonlinear function of the spatial attention mechanism in $\mathcal{S}$, \miRevise{which is implemented as} $3\times 3$ and $1\times 1$ convolutional layers with \emph{ReLU} and \emph{tanh} activation functions. $\mathbf{A}$ is the calculated attention map with the same size as $\mathbf{F}$ to score the ratio of background constituent for each pixel in $\mathbf{F}$. Thus the background constituent $\mathbf{F}_b$ is obtained by the element-wise multiplication $\odot$ between $\mathbf{A}$ and $\mathbf{F}$. It is worth mentioning that we perform such constituent separation in the deep feature space rather than on the original input image since the background and noise are not necessarily linearly separable from the input image. 

The background generator $\mathcal{G}_b$ and the noise generator $\mathcal{G}_r$ share a similar deep-masking structure, which iteratively stacks the Residual Deep-Masking Cell (presented in Section~\ref{sec:RDMC}) to perform image synthesis progressively. Long-range skip-connections are introduced between corresponding layers of the same feature size to replenish low-level features and facilitate information propagation.

The refining phase of \emph{DMGN} is designed to refine the coarse background image $B_\text{coarse}$ with the reference to the generated noise image $R$:
\begin{equation}
B_{\text{refining}} = \mathcal{G}_\text{refine} (B_{\text{coarse}}, R, I).
\label{eqn:refine}
\end{equation}
The refining generator $\mathcal{G}_\text{refine}$ \miRevise{possesses} the similar deep-masking structure as generators $\mathcal{G}_b$ and $\mathcal{G}_r$ in the coarse phase with two differences: 1) $\mathcal{G}_b$ and $\mathcal{G}_r$ first downsample the latent feature maps and then upsample them during the coarse phase, which is particularly effective in image-to-image synthesis tasks~\cite{isola2017image} due to the larger receptive field for capturing context and lower computation cost compared to the structure with fixed-size latent feature maps. Nevertheless, $\mathcal{G}_\text{refine}$ is responsible for refining the local details of $B_{\text{coarse}}$, hence its feature map size keeps constant during the refining phase; 2) the generated noise image is leveraged as the contrasting cue by the refining generator $\mathcal{G}_\text{refine}$ to facilitate the refining process, which will be explained concretely in Section~\ref{sec:refine}.

\subsection{Residual Deep-Masking Cell}
\label{sec:RDMC}
Residual Deep-Masking Cell (RDMC) is customized as the core operating unit in all three image generators of \emph{DMGN}, including $\mathcal{G}_b$, $\mathcal{G}_r$ and $\mathcal{G}_{\text{refine}}$. It models an attention mask for each pixel of feature maps using attention mechanism to control the information propagation: enhance the effective information and suppress the negative information for image synthesis. Hence, the desired image can be synthesized progressively by iteratively employing the Residual Deep-Masking Cell. 

\begin{figure}[t]
\centering
		\includegraphics[width=0.98\linewidth]{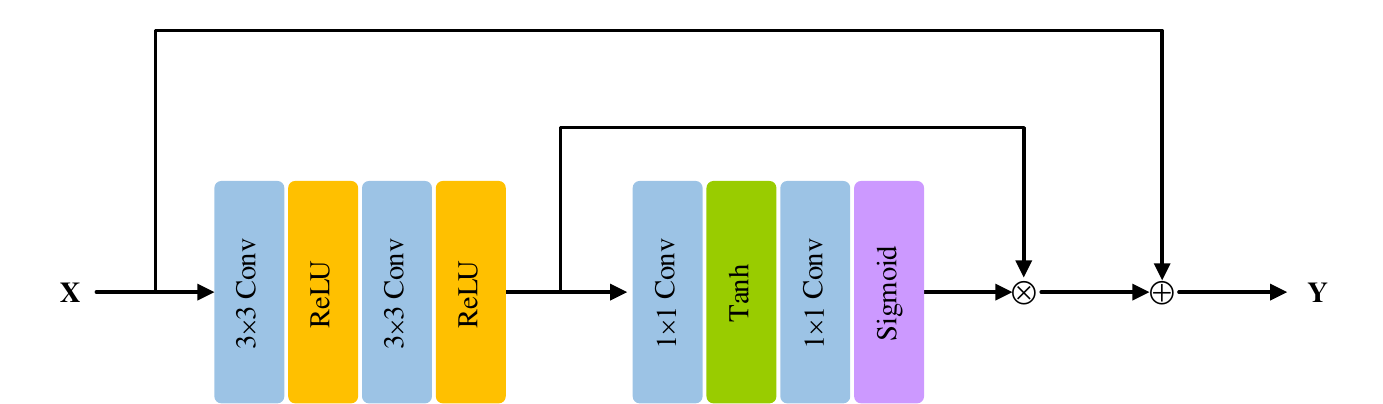}
\vspace{-4pt}
\centering
\caption{The structure of Residual Deep-Masking Cell, \miRevise{which serves as the core operating unit in the generative process.}}
\vspace{-4pt}
\label{fig:RDMC}
\end{figure}

Figure~\ref{fig:RDMC} presents the structure of proposed Residual Deep-Masking Cell (RDMC). Taking the output feature maps $\mathbf{X}$ from the previous Residual Deep-Masking Cell, two convolutional layers combining with \emph{ReLU} functions are first employed to refine the features. Then the refined feature maps $\mathcal{H}(\mathbf{X})$ are fed into an attention module, constructed by two $1 \times 1$ convolutional layers and a \emph{tanh}-function layer, to learn an mask $\mathcal{M}(\mathcal{H}(\mathbf{X}))$ with the equal size as the input feature maps. The pixel values of the mask are constrained into the interval $[0, 1]$ by a sigmoid function and measure the effectiveness of the feature maps at the corresponding pixels for image synthesis. The obtained mask are used to the control the information propagation by performing the element-wise multiplication with the refine feature maps:
\begin{equation}
    \mathbf{Y} = \mathcal{M}(\mathcal{H}(\mathbf{X})) \odot \mathcal{H}(\mathbf{X}).
    \vspace{-2pt}
\end{equation}
To avoid the potential information vanishing problem resulted from deeply stacking the Residual Deep-Masking Cell (RDMC), residual connection is applied between the input and the output of the Residual Deep-Masking Cell:
\begin{equation}
    \mathbf{Y} = \mathbf{X} + \mathcal{M}(\mathcal{H}(\mathbf{X})) \odot \mathcal{H}(\mathbf{X}),
    \label{eqn:RDMC}
    \vspace{-2pt}
\end{equation}
where $\mathbf{Y}$ is the output of current Residual Deep-Masking Cell. 

\smallskip\noindent\textbf{Physical interpretation.} In each iteration of applying Residual Deep-Masking Cell in three generators of \emph{DMGN}, the mask $\mathcal{M}(\mathcal{H}(\mathbf{X}))$ serves as an information filter to intensify the required information for image synthesis by large mask pixel values closing to 1 and filter out the irrelevant information with small pixel values closing to 0. It has greater effect in the earlier stage of the image synthesis. As the proceeding of the image synthesis, the synthesizing process is gradually finished and less information is required. Hence, more and more pixel values in the mask tend to converge to 0. The visualizations of the four masks in different periods of the background generator $\mathcal{G}_b$ in Figure~\ref{fig:intro} manifests the consistent results with above theoretical analysis. 

It should be noted that the mask of Residual Deep-Masking Cell plays opposite roles in the background generator ($\mathcal{G}_b$ or $\mathcal{G}_{\text{refine}}$) and the noise generator $\mathcal{G}_r$ due to the opposite synthesizing objectives. For instance, the mask values corresponding to the background regions in the feature maps would be high (close to 1) in background generators but low (close to 0) in a noise generator. 

\maRevise{While our proposed Residual Deep-Masking Cell (RDMC) adopts the classical attention mechanism for implementation, the major difference between our RDMC and the typical way of using attention mechanism is that RDMC is designed as the core operating module that is iteratively employed by \emph{DMGN} to perform image restoration progressively. In each iteration, RDMC performs information filtering to enhance the required information for image restoration by a residual mechanism, thus the learned attention mask (map) is used to measure the real-time \miRevise{pixel-wise} effectiveness for image restoration at current restoration stage, which \miRevise{gradually} decreases to zero for all pixels as the completion of the restoration process. By contrast, the attention mechanism is typically used to measure the relevance map of features with the objective task, e.g. the probability of being background in image restoration, which is a static evaluation value for each pixel. 
}
\subsection{Collaborative Refinement by Contrasting with Noise Layer}
\label{sec:refine}
The goal of our \emph{DMGN} is to remove noise in the input image and restore a high-quality background image. \maRevise{Since the noise and the background are two mutually exclusive constituents in a superimposed image, the obtained noise image can be used as contrasting cues for recognizing the background information. Such \miRevise{contrastive} strategy to circumvent the challenges of image restoration is also widely adopted by a type of traditional methods~\cite{chen2013generalized,li2016rain,zhu2017joint}, which leverage prior knowledge about noise to model the noise layer and then obtain the background image by contrasting with the noise layer.} Similarly, we leverage the generated noise image as contrasting cues to facilitate the refinement of generated coarse background image in the refining phase, as shown in Equation~\ref{eqn:refine}.


We propose a two-pronged strategy to make use of the generated noise image in the refinement of the coarse background image: 1) \miRevise{the noise image is incorporated as a regularization for the source information for the refining generator $\mathcal{G}_{\text{refine}}$; 2) the noise image is utilized to regularize the mask values of Residual Deep-Masking Cell (RDMC) iteratively deployed in the generator $\mathcal{G}_{\text{refine}}$.} These two mechanisms are jointly employed to fully leverage the generated noise image and thus collaboratively aid the refining generator to refine the background image. 

\smallskip\noindent\textbf{Regularize the source information for the refining generator $\mathcal{G}_{\text{refine}}$.}
\label{sec:rsi}
 The noise image $R$ generated in the coarse phase is incorporated into the source information for the refining generator $\mathcal{G}_{\text{refine}}$ to provide contrasting cues. Besides, the separated background content $\mathbf{F}_b$ (indicated in Equation~\ref{eqn:coarse}) from the input image by the separator $\mathcal{S}$ in the coarse phase is also incorporated for the source information of $\mathcal{G}_{\text{refine}}$ as positive cues. Providing both such contrasting cues and positive cues, the refining generator $\mathcal{G}_{\text{refine}}$ is expected to learn the background features more precisely and thereby generating a high-quality background image. 

 Specifically, we propose an Adapted Channel-wise Attention module (\emph{ACA}) to fuse the noise content and the separated background content $\mathbf{F}_b$:
 \begin{equation}
 \begin{split}
     &\mathbf{E}_b = \emph{ACA}(\mathbf{F}_b), \\
     &\mathbf{E}_r = - \emph{ACA}(\mathcal{F}(R)), \\
     &\mathbf{E} = \text{Concat}(\mathbf{E}_b, \mathbf{E}_r, \mathcal{F}(B_{\text{coarse}})).
\end{split}
 \end{equation}
 Herein, $\mathbf{E}_b$ and $\mathbf{E}_r$ are the transformed feature maps for background and noise respectively by \emph{ACA}. Note that $\mathbf{E}_r$ is negated since it serves as contrasting cues, which is opposite to the positive cues $\mathbf{E}_b$. They are concatenated with the coarse background content to form the source information $\mathbf{E}$ for $\mathcal{G}_{\text{refine}}$. Both the noise image $R$ and the coarse background image $B_{\text{coarse}}$ are first projected into the same feature space by the feature-learning head $\mathcal{F}$ before information fusion. As shown in Figure~\ref{fig:model}, \emph{ACA} is designed similar to the Squeeze-and-Excitation block in SENet~\cite{hu2018squeeze} with a major difference: the pixel value is normalized by a \emph{sigmoid}
 function to obtain a sharper pixel distribution for each channel of feature map. Formally, given feature maps $\mathbf{Feat}$ as input for \emph{ACA}, the $i$-th channel in $\mathbf{Feat}$ is normalized by:
 \begin{equation}
 \mathbf{Feat}'_i = \sigma (w_i \cdot f_{\text{conv}}(\mathbf{Feat}_i)) \odot f_{\text{conv}}(\mathbf{Feat}_i).
 \end{equation}
 Here the parameter $w_i$ is learned to tune the sharpness of the \emph{sigmoid} function and $f_{\text{conv}}$ is the transformation function \miRevise{implemented} by a $5\times 5$ convolution layer and a \emph{ReLU} layer for downsampling.  Then the normalized feature map is fed into a Squeeze-and-Excitation block to compute the \miRevise{channel-wise} attention and obtain the final output of \emph{ACA}:
 \begin{equation}
 \begin{split}
      &\mathbf{a} = \text{SE-block}(\mathbf{Feat}'), \\
      &\emph{ACA}(\mathbf{Feat}_i) = \mathbf{a}_i \cdot \mathbf{Feat}_i, i \in \{1, \dots, C\},
\end{split}
\vspace{-3pt}
 \end{equation}
where $\mathbf{a}$ is the calculated channel attention and $C$ is the number of channel in $\mathbf{Feat}$. We normalize each channel of feature maps to have a sharper pixel distribution and thus assign higher weights to larger-value pixels when modeling the channel attention. 

\noindent\textbf{Regularize the mask of Residual Deep-Masking Cell.}
The second mechanism of leveraging the generated noise image for refining the background image in our \emph{DMGN} is to regularize the mask values of Residual Deep-Masking Cell involved in the refining generator $\mathcal{G}_{\text{refine}}$. Specifically, we employ an attention model to estimate the distribution map of noise in the input image based on the obtained noise image:
\begin{equation}
     \mathbf{A}_r = \sigma (f'_{att}[\mathbf{\mathcal{F}(I), \mathcal{F}(R)}]), 
     \vspace{-3pt}
\end{equation}
where $f'_{att}$ is the transformation function by the attention model and $\mathbf{A}_r$ is the calculated attention map for estimating the noise distribution in the input image. Here the attention model shares the similar structure as the attention model in the separator $\mathcal{S}$ except for the concatenation layer, as shown in Figure~\ref{fig:model}.

\begin{figure}[t]
\centering
	\includegraphics[width=0.9\linewidth]{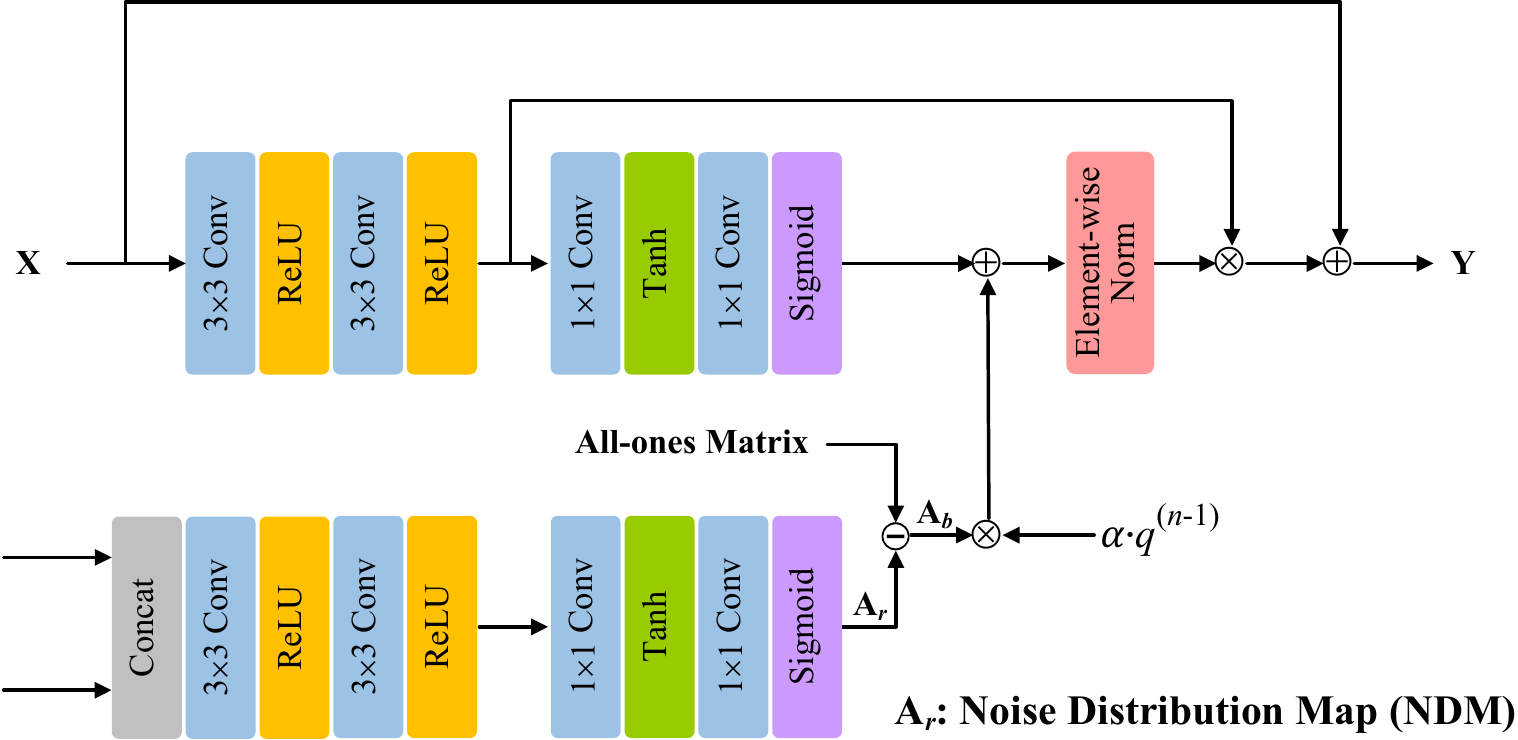}
	\vspace{-4pt}
\caption{The structure of Residual Deep-Masking Cell regularized by contrasting with noise.}
\vspace{-4pt}
\label{fig:RDMC2}
\end{figure}

The obtained noise distribution map (NDM) $\mathbf{A}_r$ is utilized to infer the background distribution map by $\mathbf{A}_b = 1 - \mathbf{A}_r$. Then we use it to regularize the mask values (in Equation~\ref{eqn:RDMC}) of $n$-th Residual Deep-Masking Cell in the refining generator $\mathcal{G}_{\text{refine}}$ by fusing it with the learned mask value by itself $\mathcal{M}(\mathcal{H}(\mathbf{X}))$:
\begin{equation}
    \mathcal{M}(\mathbf{X})  = \frac{\mathcal{M}(\mathcal{H}(\mathbf{X})) + \alpha \cdot q^{(n-1)} \cdot \mathbf{A}_b}{1+\alpha \cdot q^{(n-1)}}.
    \label{eqn:regu2}
\end{equation}
Here $\alpha \in [0,1]$ is a hyper-parameter to balance between two terms. $q \in [0,1]$ is a weight-decaying factor to diminish the effect of $\mathbf{A}_b$ on the mask value as the proceeding of the refining phase.
\maRevise{Owning to the fusion mechanism designed in Equation~\ref{eqn:regu2}, $\mathbf{A}_b$ is learned to indicate the background distribution map which is consistent with the mask $\mathcal{M}(\mathcal{H}(\mathbf{X}))$  under the supervision of final loss functions. Accordingly, $\mathbf{A}_r$ is learned to reflect the noise distribution map.}
Figure~\ref{fig:RDMC2} illustrates the structure of the Residual Deep-Masking Cell regularized by contrasting with the noise image.

The rationale behind such regularization is that the noise distribution map $\mathbf{A}_r$ and the background distribution map $\mathbf{A}_b$ are estimated explicitly by comparing between input image and the generated noise image under the supervision of final loss functions, while the mask values $\mathcal{M}(\mathcal{H}(\mathbf{X}))$ of Residual Deep-Masking Cell is learned solely under the supervision on the generated background image. Hence they are derived from different knowledge.

\begin{figure*}[!t]
  \centering 
  \begin{minipage}[b]{0.93\linewidth}
  \includegraphics[width=\linewidth]{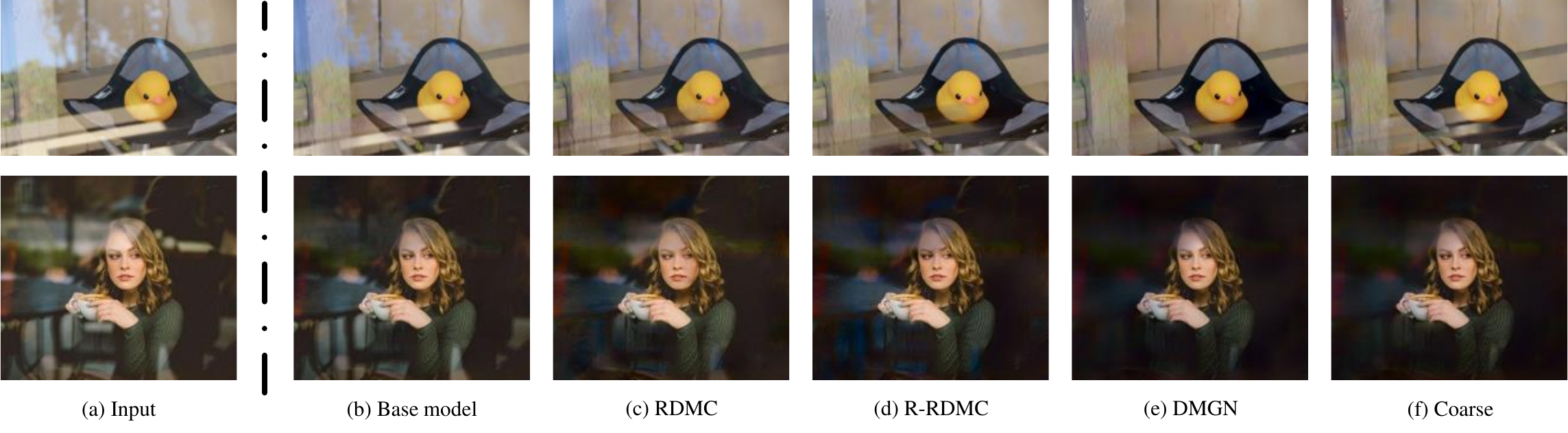}\vspace{4pt}
  \end{minipage}
  \vspace{-4pt}
\caption{Visualization of the restored background images by five variants of our \emph{DMGN} for image reflection removal on two randomly selected samples from test set.}
  \vspace{-4pt}
\label{fig:ablation_qualitative}
\end{figure*}

\subsection{End-to-end Parameter Learning by Joint Supervision}    
We optimize the whole model of \emph{DMGN} in an end-to-end manner by performing supervised learning on both the coarse phase and the refining phase. Specifically, we employ three types of loss functions to train our \emph{DMGN}:
\begin{itemize}[leftmargin =*]
\item \textbf{Pixel-wise L1 reconstruction Loss} which focuses on pixel-level measurement: 
\begin{equation}
    \mathcal{L}_{L1}  = \|Y - Y_{GT} \|_1,
\end{equation}
where $Y$ and $Y_{GT}$ are the generated image and the corresponding ground-truth image respectively. 

\item \textbf{Perceptual loss}~\cite{johnson2016perceptual} \miRevise{that is proposed to minimized the semantic difference between synthesized image and the groundtruth image} in deep feature space:
\begin{equation}
    \mathcal{L}_{p} (Y, Y_{GT}) = \sum_{l=1}^L \frac{1}{C_l  H_l W_l} \| f^l_\text{vgg}(Y) - f^l_\text{vgg}(Y_{GT})\|_1,
\end{equation}
where $f^l_\text{vgg}(Y)$ and $f^l_\text{vgg}(Y_{GT})$ are the extracted feature maps (with dimensions $C_l \times H_l \times W_l$) for the generated image $Y$ and the ground-truth image $Y_{GT}$ respectively from the $l$-th convolutional layer of the pre-trained VGG-19 network~\cite{simonyan2014very}.

\noindent \item \textbf{Conditional Adversarial loss}~\cite{mirza2014conditional} \miRevise{which encourages} the generated background image $B$ to be as realistic as the ground-truth background image $B_{\text{GT}}$:
\begin{equation}
    \mathcal{L}_{\text{adv}}  = -\mathbb{E}_{B\backsim \mathbb{P}_\text{DMGN}}[D(B, I)],
\end{equation}
where $I$ is the input image with noise and the Discriminator $D$ is trained by:
\begin{equation}
\resizebox{0.9\linewidth}{!}{$
    \quad \mathcal{L}_{D} = -\mathbb{E}_{B_{\text{GT}}\backsim \mathbb{P}_\text{data}}[D(B_{\text{GT}}, I)]+\mathbb{E}_{B\backsim \mathbb{P}_\text{DMGN}}[D(B, I)].
    $}
\end{equation}
We adopt the same structure of discriminator as \emph{pix2pix}~\cite{isola2017image}.
\end{itemize}

The refining phase in our \emph{DMGN} employs all these three types of loss functions:
\begin{equation}
    \mathcal{L}^{\text{r}} = \alpha^{\text{r}}_{L1}\mathcal{L}_{L1}^{\text{r}} +  \alpha^{\text{r}}_p\mathcal{L}_{p}^{\text{r}}+  \alpha^{\text{r}}_{\text{adv}}\mathcal{L}_{\text{adv}}^{\text{r}},
\end{equation}
where $\alpha^{\text{r}}_{L1}$, $\alpha^{\text{r}}_p$ and $\alpha^{\text{r}}_{\text{adv}}$ are the hyper-parameters to balance between different losses. 

The adversarial loss is not applied to the coarse phase of \emph{DMGN} since \miRevise{the quality of the coarsely generated background image is not adequate} for the discrimination from the ground-truth image. Hence, the coarse phase is supervised by the L1 reconstruction loss and the perceptual loss:
\begin{equation}
    \mathcal{L}^{\text{c}} = \alpha^{\text{c}}_{L1}\mathcal{L}_{L1}^{\text{c}} +  \alpha^{\text{c}}_p\mathcal{L}_{p}^{\text{c}}.
\end{equation}
Consequently, the coarse phase and the refining phase of \emph{DMGN} are jointly optimized by:
\begin{equation}
    \mathcal{L} = \mathcal{L}^{\text{r}} + \mathcal{L}^{\text{c}}.
\end{equation}

%% file: experiments.tex
To evaluate the generalization and robustness of our Deep-Masking Generative Network (\emph{DMGN}) across different tasks with diverse noise patterns, we conduct experiments on three image background restoration tasks: (1) image reflection removal, (2) image rain streak removal and (3) image dehazing. Besides, we also perform ablation study on the task of image reflection removal to investigate the effect of each proposed functional technique in our \emph{DMGN}.

\subsection{Experimental Setup \maRevise{Shared across  Experiments}}
\smallskip\noindent\textbf{Evaluation Metrics.}
We adopt two commonly used metrics to quantitatively measure the quality of the generated background images in our experiments: \emph{PSNR} and \emph{SSIM}. \emph{PSNR} computes the peak signal-to-noise ratio in decibels between two images while \emph{SSIM} measures the perceptual similarity between two images. Higher value of \emph{PSNR} or \emph{SSIM} implies higher quality of the generated background image.

Additionally, we also perform qualitative evaluation by visually \miRevise{evaluating} the restored background results of randomly selected test samples by different models for each task in experiments.

As a complement to the standard evaluation metrics, we further perform human evaluation to compare our model to the state-of-the-art methods. \miRevise{Considering the substantial amount of labor workload, we only perform such human evaluation for image reflection removal.}

\smallskip\noindent\textbf{Implementation Details.}
We implement our \emph{DMGN} in distribution mode with 4 Titan RTX GPUs under Pytorch framework. Adam~\cite{kingma2014adam} is employed for gradient descent optimization with batch size set to be 4. The initial learning rate is set to be $2\times 10^{-4}$ and the training process takes maximally 100 epochs. By tunning on a held-out validation set, the hyper-parameters $\alpha^{\text{r}}_{L1}$, $\alpha^{\text{c}}_{L1}$, $\alpha^{\text{r}}_p$, $\alpha^{\text{c}}_p$ and $\alpha^{\text{r}}_{\text{adv}}$ in the loss functions are set to be 1, 0.5, 0.1, 0.05, 0.01 respectively. Random flipping, random cropping and resizing are used for data augmentation. 

\maRevise{To have fair comparisons between different methods for each image restoration task, we ensure that all methods are optimized on the same training set and evaluated on the same test set following the same optimization and evaluation protocols. 
}

\begin{figure}[t]
\centering
	\includegraphics[width=0.98\linewidth]{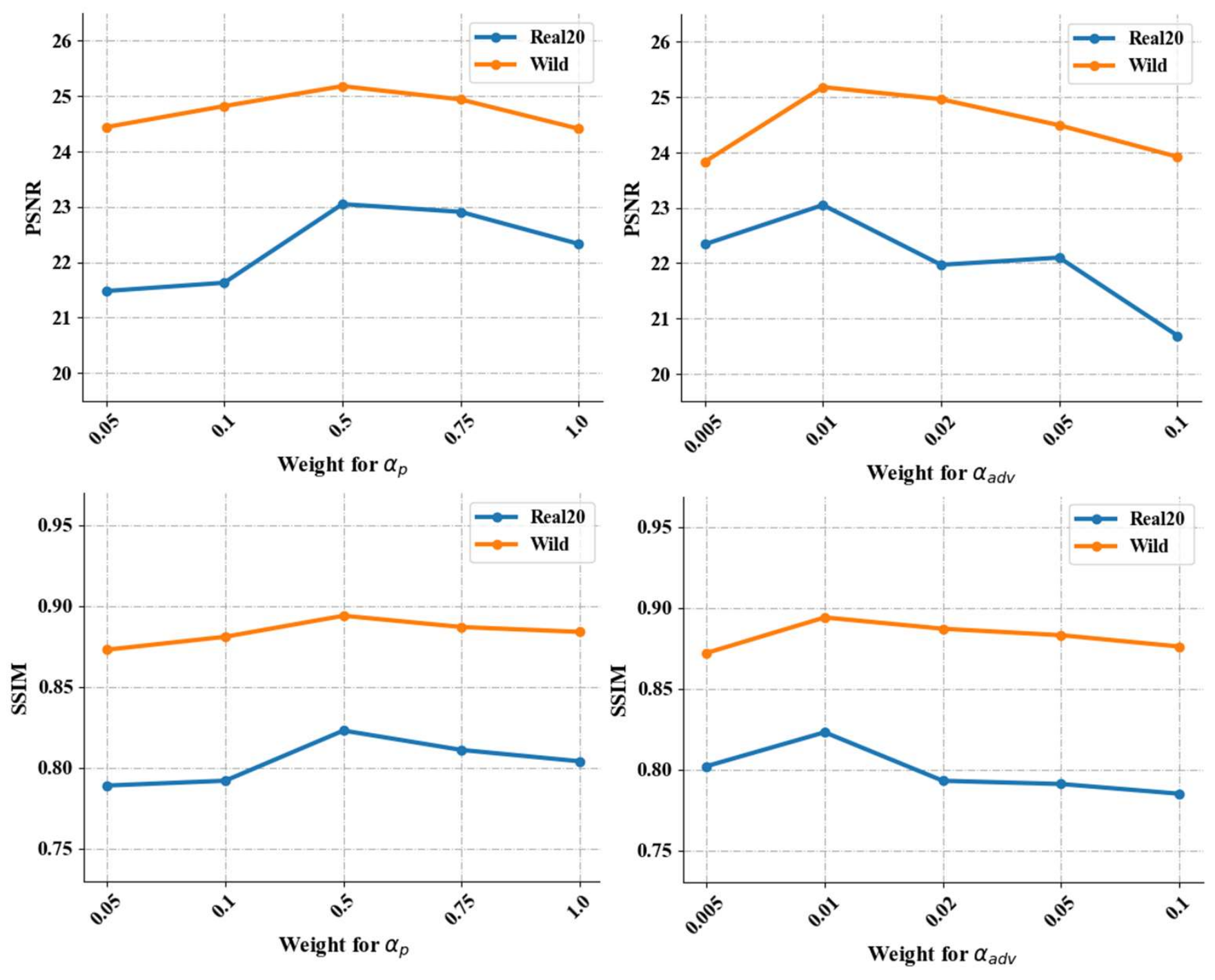}
\caption{\maRevise{Performance of our \emph{DMGN} as a function of loss weight $\alpha_p$ and $\alpha_{adv}$ when setting $\alpha_{L1}$ as 1.}}
\label{fig:wegihts}
\end{figure}

\subsection{Experiments on Image Reflection Removal}
\smallskip\noindent\textbf{Datasets.} Following PLNet~\cite{zhang2018single}, the training data in our experiments consists of two parts: 1) synthetic data, which is synthesized from randomly selected pairs of images from Flickr in the same way as PLNet; 2) Real data, which is collected in PLNet and is composed of 2400 pairs of patches randomly cropped from 90 training images. 

We perform quantitative evaluation on four real-world datasets: 1) Real20~\cite{zhang2018single} including 20 real-world image pairs across various scenes. 2) Solid dataset~\cite{wan2017benchmarking} consisting of 200 triplets of images depicting indoor solid object scenes; 3) Postcard dataset~\cite{wan2017benchmarking}, which is \miRevise{a} challenging dataset containing 199 triplets of images obtained from postcards; 4) Wild dataset~\cite{wan2017benchmarking} that contains 55 triplets of images about wild scenes. Besides, we also utilize two more real-world datasets for qualitative evaluation: real-world images collected by CEILNet~\cite{fan2017generic} and ERRNet~\cite{wei2019single} respectively. Since the ground-truth background images are not provided in these two datasets, they cannot be used for quantitative evaluation.

\input{histogram.tex}

\begin{figure*}[!t]
  \centering 
  \begin{minipage}[b]{0.99\linewidth}
  \includegraphics[width=\linewidth]{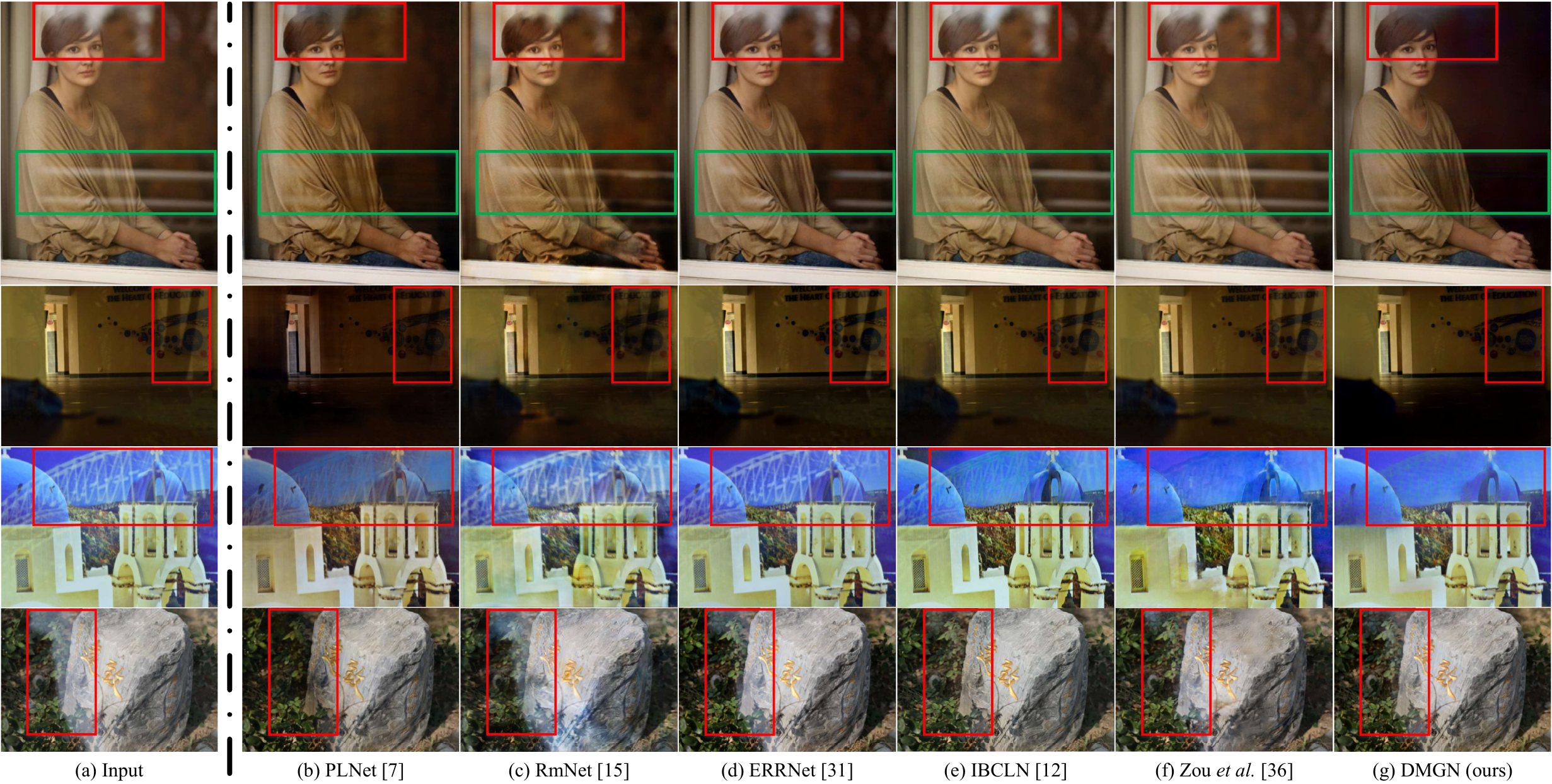}\vspace{4pt}
  \end{minipage}
\caption{Visualization of restored background images by seven models for image reflection removal on four randomly selected samples from test set. Our model is able to restore cleaner and higher-quality background image than other methods, particularly in the regions indicated by bounding boxes. Best viewed in zoom-in mode.}
  \vspace{-4pt}
\label{fig:qualitative}
\end{figure*}

\subsubsection{Ablation Study}
We first conduct experiments to investigate the effect of each proposed functional technique in our \emph{DMGN}. To this end, we perform ablation experiments on five variants of our \emph{DMGN} as well as the perceptual loss $\mathcal{L}_p$ and conditional adversarial loss $\mathcal{L}_{adv}$:
\begin{itemize}[leftmargin=*]
    \item \textbf{Base model}, which only employs the coarse-to-fine generative framework without generating the reflection image in the coarse phase. Thus, no reflection image is leveraged as contrasting cues in the refining phase. Besides, the proposed Residual Deep-Masking Cell is not applied yet, namely, the plain convolutional layer is utilized as the iterative cell structure for image generators. 
    
    \item\textbf{RDMC}, which employs the proposed \textbf{R}esidual \textbf{D}eep-\textbf{M}asking \textbf{C}ell to replace the plain CNN structure in the base model as the iterative operating unit for image synthesis. 
    
    \item\textbf{R-RDMC}, which augments the model by generating the reflection image in the coarse phase and leverages the reflection image to regularize the mask values of the residual deep-masking cells in the refining phase, as described in Section~\ref{sec:refine}.
    
    \item\textbf{DMGN}, which further leverages the generated reflection image to regularize the source information for the refining generator based on R-RDMC, as described in Section~\ref{sec:rsi}. The resulting model is our intact \emph{DMGN}. 
    
    \item\textbf{Coarse}, which corresponds to the coarse phase in our \emph{DMGN}.
    \item\maRevise{\textbf{w/o $\mathcal{L}_\text{adv}$}, which does not utilize adversarial loss in the refining phase.}
    \item\maRevise{\textbf{w/o $\mathcal{L}_\text{p}$}, in which the perceptual loss in not used \miRevise{during training}.}
    
\end{itemize}

Table~\ref{tab:ablation} presents the experimental results of seven variants of \emph{DMGN} on Real20 and Wild datasets, in terms of both \emph{PSNR} and \emph{SSIM}. 

\noindent\textbf{Effect of Residual Deep-Masking Cell}.
The large performance gaps between \textbf{Base model} and \textbf{RDMC} in terms of both \emph{PSNR} and \emph{SSIM} demonstrate the remarkable advantage of the proposed Residual Deep-Masking Cell (RDMC) compared to plain convolutional layer for the task of image reflection removal. \miRevise{The crux of synthesizing either the background image or the reflection image} is to recognize the useful information and filter out the negative information from the input image during synthesis, which is the goal of designing RDMC. However, this is hardly achieved by plain convolutional layer.

\noindent\textbf{Effect of regularizing the RDMCs in the refining generator by contrasting with reflection}.
The performance comparison between \textbf{RDMC} and \textbf{R-RDMC} manifests the benefit of regularizing the RDMCs in the refining generator by leveraging the generated reflection image. The reflection image is used to predict a reflection distribution map as explicit contrasting cues to \miRevise{refine} the mask values of RDMCs involved in the refining generator.

\input{table1.tex}

\noindent\textbf{Effect of regularizing the source information for the refining generator by contrasting with reflection}.
Comparing the performance of \textbf{R-RDMC} to \textbf{DMGN} in Figure~\ref{tab:ablation} , the performance in both \emph{PSNR} and \emph{SSIM} is further boosted by regularizing the source information for the refining generator by contrasting with the generated reflection image. It implies that two regularizing ways (performed on RDMC or source information) based on the generated reflection image contribute to the performance \miRevise{in their own way, i.e.,their effects are not identical}. 

\noindent\textbf{Effect of the refining phase}.
We also report the performance of the variant \textbf{Coarse} in ablation experiments in Table~\ref{tab:ablation} to investigate the effect of the refining phase in our \emph{DMGN}. The performance on both Real20 and Wild datasets are improved substantially by the refining phase, which reveals the significant effectiveness of the refining phase in our \emph{DMGN}.

\noindent\miRevise{\textbf{Qualitative ablation evaluation.}} Figure~\ref{fig:ablation_qualitative} presents two examples for qualitative comparison, in which the restored background images by five variants are visualized. 
It shows that the quality of background restoration becomes increasingly better as the augment of our \emph{DMGN} with different functional techniques.

\noindent\maRevise{\textbf{Effect of each loss}. Table~\ref{tab:ablation} also presents the performance of our model without the perceptual loss $\mathcal{L}_p$ or without the adversarial loss $\mathcal{L}_{adv}$. We observe that both losses yield notable performance improvement while $\mathcal{L}_{p}$ has relatively larger contribution. To have more insight into the effect of two losses, Figure~\ref{fig:wegihts} shows the performance of our model in terms of \emph{PSNR} and \emph{SSIM} as a function of loss weights $\alpha_p$ and $\alpha_{adv}$ when setting the weight $\mathcal{\alpha}_{L1}$ as 1.}

\begin{figure*}[!t]
  \centering 

\subfigure[Input]{
\begin{minipage}[b]{0.14\linewidth} 
      \centering
      \vspace{40pt}
      \includegraphics[width=\linewidth]{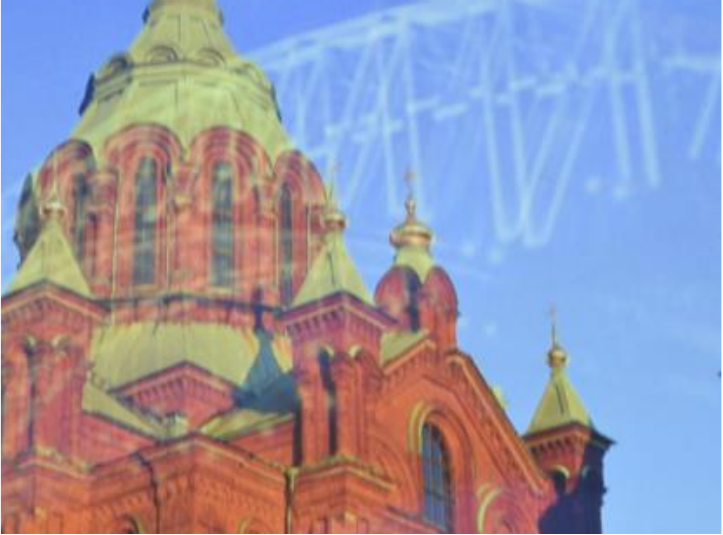}
      \vspace{20pt}
    \end{minipage}
\hfill
}
\subfigure[BDN~\cite{yang2018seeing}]{
\begin{minipage}[b]{0.14\linewidth} 
      \centering
      \includegraphics[width=\linewidth]{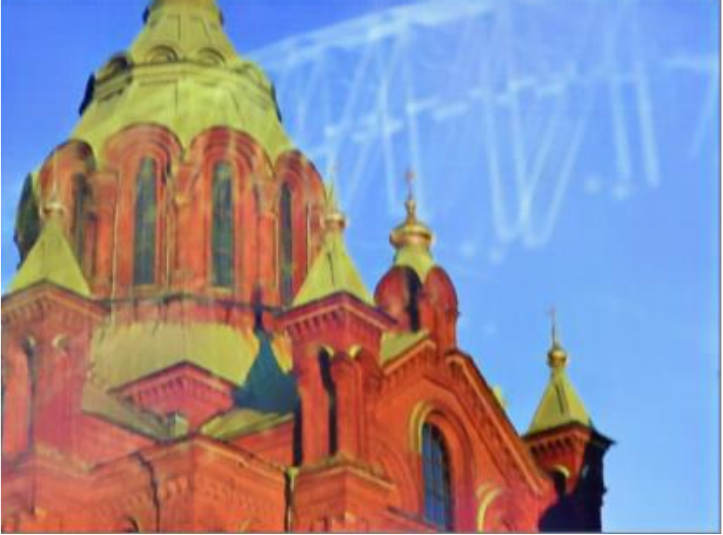}\vspace{4pt}
      \includegraphics[width=\linewidth]{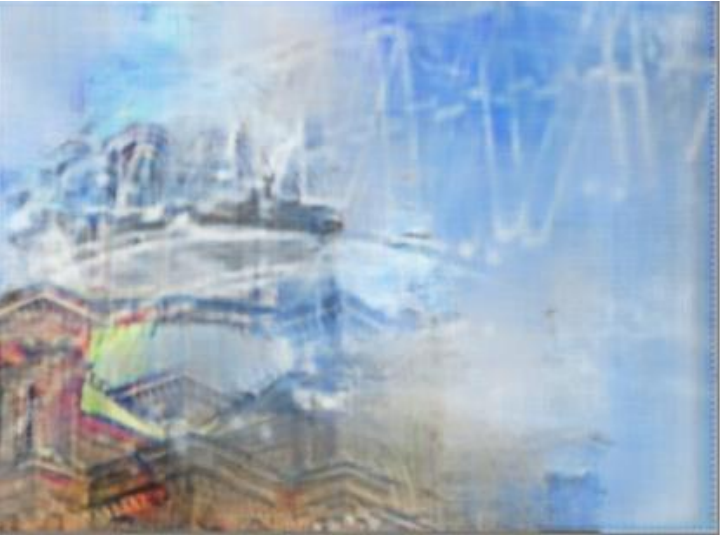}
    \end{minipage}
\hfill
}
\subfigure[PLNet~\cite{zhang2018single}]{
\begin{minipage}[b]{0.14\linewidth} 
      \centering
      \includegraphics[width=\linewidth]{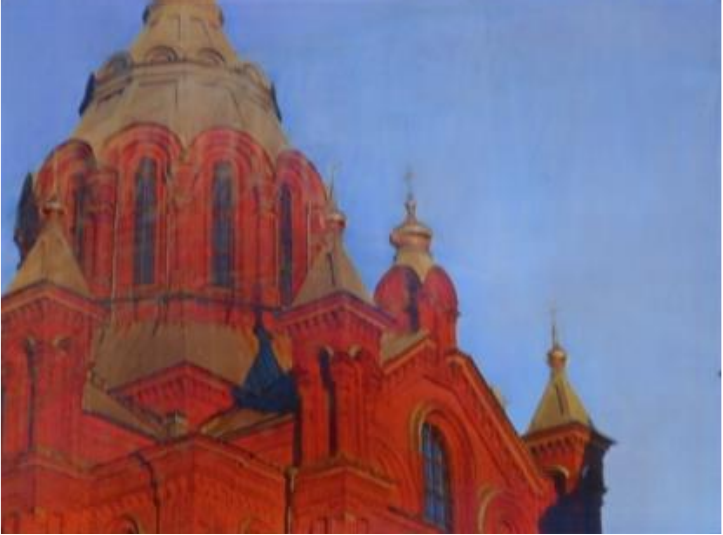}\vspace{4pt}
      \includegraphics[width=\linewidth]{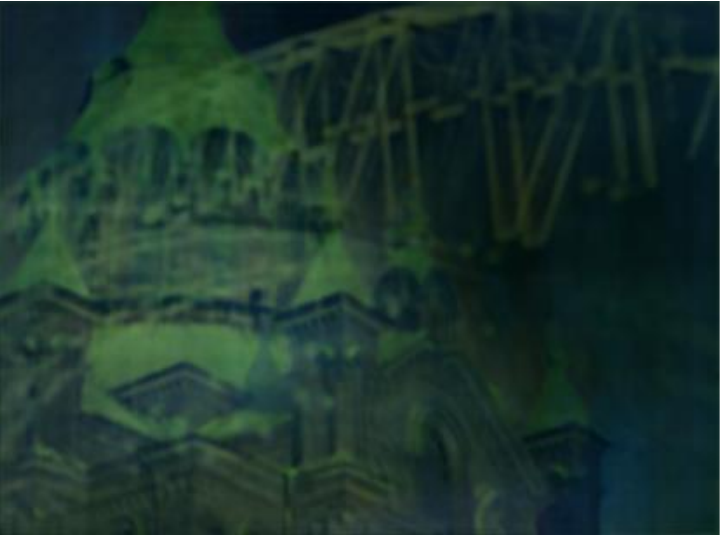}
    \end{minipage}
\hfill
}
\subfigure[RmNet~\cite{wen2019single}]{
\begin{minipage}[b]{0.14\linewidth} 
      \centering
      \includegraphics[width=\linewidth]{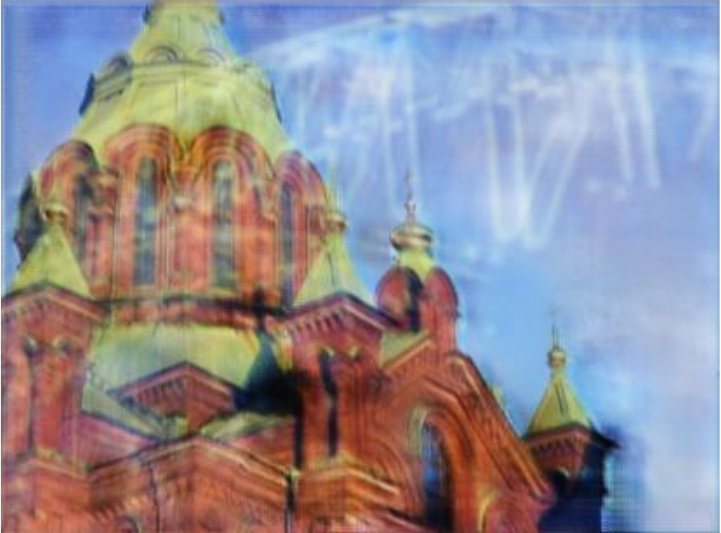}\vspace{4pt}
      \includegraphics[width=\linewidth]{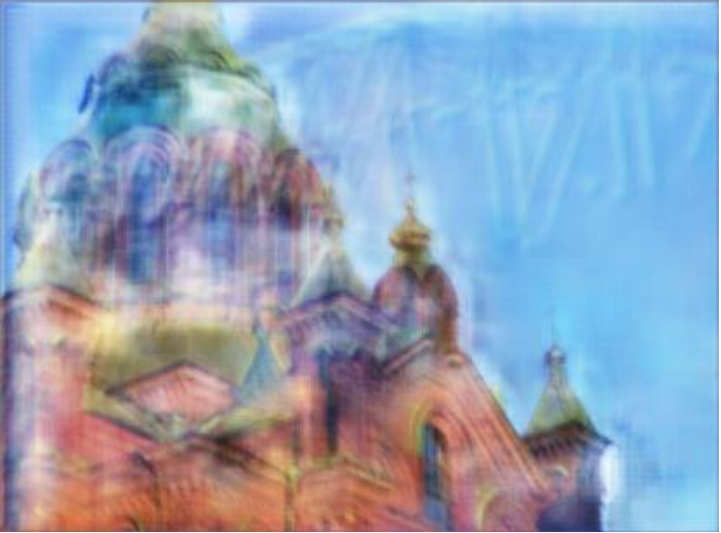}
    \end{minipage}
\hfill
}
\subfigure[DMGN (ours)]{
\begin{minipage}[b]{0.14\linewidth} 
      \centering
      \includegraphics[width=\linewidth]{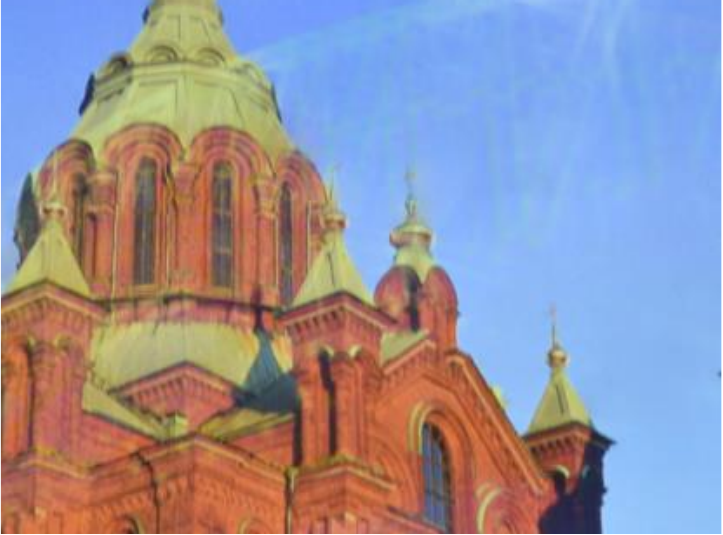}\vspace{4pt}
      \includegraphics[width=\linewidth]{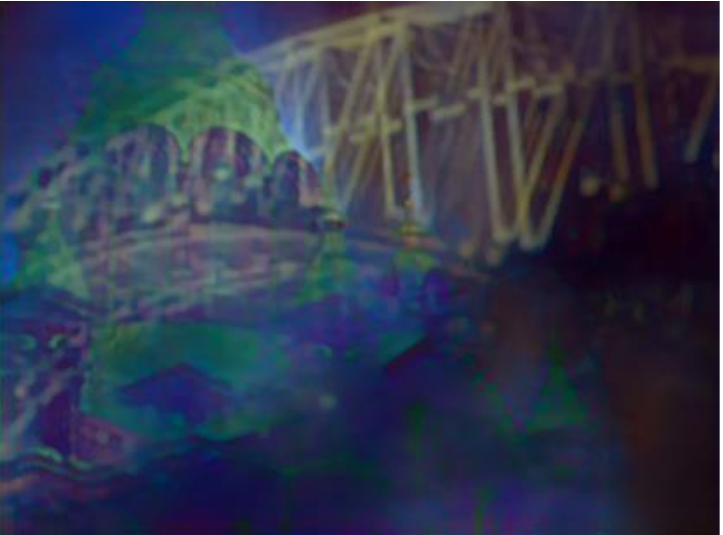}
    \end{minipage}
\hfill
}
\subfigure[GroundTruth]{
\begin{minipage}[b]{0.14\linewidth} 
      \centering
      \includegraphics[width=\linewidth]{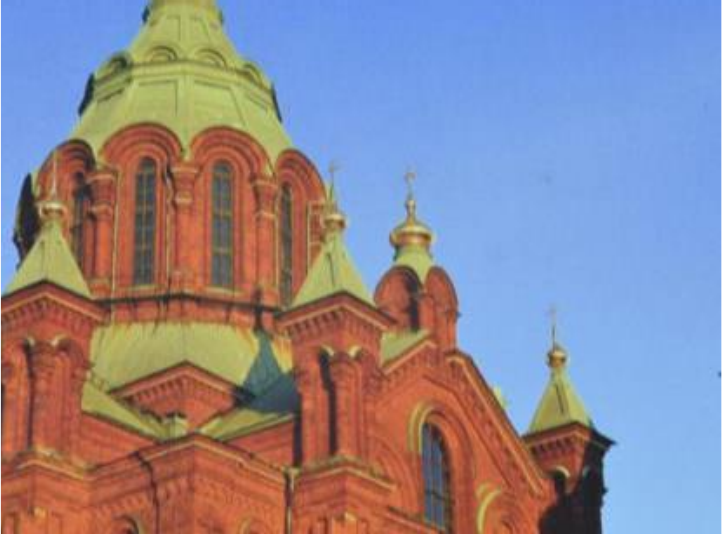}\vspace{4pt}
      \includegraphics[width=\linewidth]{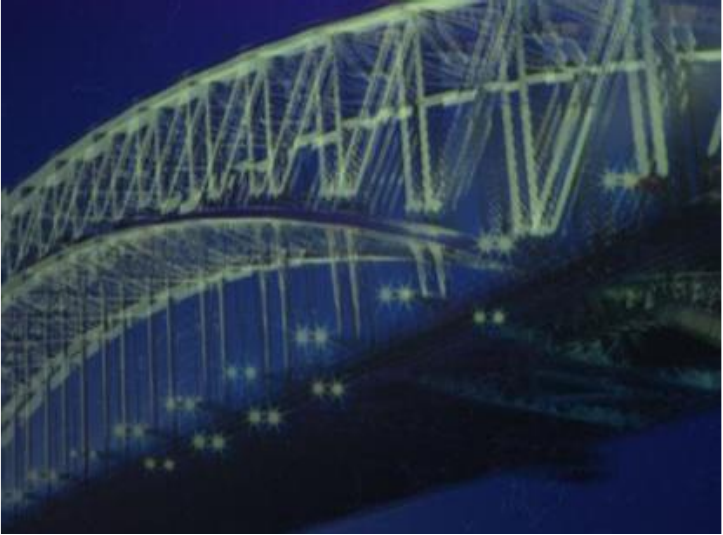}
    \end{minipage}
\hfill
}
  \vspace{-4pt}
\caption{Visualization of both the restored background images (top row) and the generated reflection images (bottom row) by four models for image reflection removal on a randomly selected sample from test set. The ground truth is also provided. Our model generates more precise reflection image than other models, which leads to higher-quality background image.}
\vspace{-4pt}
\label{fig:reflection}
\end{figure*}

\subsubsection{Comparison with State-of-the-art Methods}
Next we conduct experiments to compare our model with state-of-the-art methods for image reflection removal on four datasets including Real20, Solid, Postcard and Wild. In particular, we compare with two groups of models: ad hoc methods which are specifically designed for one specific task and unified methods which is feasible \miRevise{for various kinds of image restoration tasks. Concretely, our model is compared with} 1) \textbf{BDN}~\cite{yang2018seeing}: bidirectional network which synthesizes the background image and the reflection image alternately to improve the quality of restoration; 2) \textbf{PLNet}~\cite{zhang2018single}, which first proposes to apply perceptual loss to minimize the semantical distance between restored background image and the ground-truth for image reflection removal; 3) \textbf{RmNet}~\cite{wen2019single}, which designs a synthesis network to generate realistic images with reflection for data augmentation; 4) \textbf{ERRNet}~\cite{wei2019single}, which focuses on exploiting the misaligned training data and leveraging the multi-scale context; 5) \textbf{YW19}~\cite{yang2019fast}, which proposes an efficient convex model without the need for training based on the assumption that the sharp edges tend to appear in the background.

\smallskip\noindent\textbf{Quantitative Evaluation.}
\label{sec:reflection_exp}
Table~\ref{tab:comparison} presents the experimental results of different models for image reflection removal on four datasets in terms of both \emph{PSNR} and \emph{SSIM}. \maRevise{Our \emph{DMGN} achieves the best performance in terms of both metrics on all four datasets \miRevise{except the Solid and Wild datasets w.r.t. \emph{PSNR} (the second place)}, which indicates the effectiveness and robustness of our method across different datasets.} It is reasonable that the performance of \emph{YW19} is inferior to other methods in that it performs inference directly without learnable parameters to be trained and thus its advantage is high efficiency. It is worth mentioning that \emph{BDN} and \emph{RmNet} adopts the similar strategy as our \emph{DMGN}, i.e., leveraging the generated reflection image to facilitate the background synthesis. \miRevise{However,} \emph{BDN} just concatenates the reflection image to the input image in a brute-force way and leave the model to learn the cues by itself, which turns out to have little effect on the performance. \emph{RmNet} seeks to reconstruct the input image relying on the generated background image and the reflection image, and optimize the model by the reconstruction loss, which proves not as effective as our \emph{DMGN}. In contrast, our \emph{DMGN} models the generated reflection image explicitly as contrasting cues for collaborative background refinement by regularizing both the source information and the Residual Deep-Masking Cells in the refining generator. 

\emph{ERRNet} performs better than other models except for our \emph{DMGN}. The good performance of \emph{ERRNet} is resulted from its two techniques: \miRevise{the proposed} alignment-invariant loss to exploit the misaligned training data and the multi-scale context exploration module. These two techniques can be readily integrated into our \emph{DMGN}, leading to a more powerful system for image reflection removal. 

Note that Zou \emph{et al.}~\cite{Zou_2020_CVPR}, who has the similar motivation as our model, aims to propose a unified solution for separating the superimposed images based on multi-level adversarial supervision. We conduct experiments on same datasets with its official source code provided online. The experimental results in Table~\ref{tab:comparison} shows that our model distinctly outperforms it on all four datasets except in \emph{PSNR} on Wild dataset, which indicates the superiority of our model over it.

\maRevise{To investigate whether the performance superiority of our model over other methods is resulted from the advantages of model design or more model complexity, we compare the model complexity of our model with two most recent state-of-the-art methods for image reflection removal including \emph{ERRNet} and \emph{IBCLN} in Table~\ref{tab:efficiency}. The results show that our model has much less GFLOPS and inference time than other two methods, which resolves the doubt and reveals that the distinct performance gain of our model over these two methods are not contributed by more model complexity.
}

\smallskip\noindent\textbf{Qualitative Evaluation.}
To gain more insight into the \miRevise{performance} of image reflection removal by different models, we randomly select four samples from the test set and visualize the generated background images by these models in Figure~\ref{fig:qualitative}. It qualitatively manifests our \emph{DMGN} performs best among all models, particularly in the regions indicated by bounding boxes. Our \emph{DMGN} is able to remove most of reflection in the input image while retaining the background content. However, other methods either cannot remove the reflection clearly or remove some background content falsely (e.g, the darker background images restored by \emph{PLNet} due to excessive reflection removal). Note that \emph{ERRNet} performs inferior to our model and \emph{PLNet}, although it achieves much higher quantitative performance than \emph{PLNet} in terms of \emph{PSNR} and \emph{SSIM} shown in Table~\ref{tab:comparison}.
It somewhat reveals that these quantitative metrics are not absolutely impartial. 

\input{human_table.tex}

Since the generated reflection image is leveraged to facilitate the background refinement in our \emph{DMGN}, the quality of the generated reflection image is crucial to the refinement. Thus, we also randomly select a \miRevise{test} example and visualize the generated reflection image by our model as well as the ground-truth reflection image in Figure~\ref{fig:reflection}. Besides, we also present the visualization of generated reflection images by \emph{BDN}, \emph{PLNet} and \emph{RmNet}, which also leverage the generated reflection image to improve the quality of the background image. It shows that our model generates more precise reflection image than other three models. Nevertheless, the quality of the generated reflection images are relatively worse than the background image due to the major optimizing objective of restoring background images.

\smallskip\noindent\textbf{Human Evaluation.} Both \emph{PSNR} and \emph{SSIM} have their bias for quality evaluation of restored background images. \maRevise{As a complement to the standard evaluation metrics, we further perform human evaluation to compare our model with other top-three most powerful methods for reflection removal including  \emph{ERRNet}~\cite{wei2019single}, \emph{IBCLN}~\cite{li2019single} and Zou \emph{et al.}~\cite{Zou_2020_CVPR}. We randomly select 50 test samples and present the generated background images by our model and other three methods to 50 human subjects for manual ranking of restoring quality \miRevise{(between different methods for each sample)}. As shown in Table~\ref{tab:human}, among total 50$\times$50 = 2500 rankings, our model achieves $53.2\%$ of 2500 votes which is much more than other three models. Besides, we aggregate evaluation results of all subjects for each sample, our model wins on 42 test samples and other three models win on totally 6 test samples (the left 2 samples have equal votes for two or more than two methods).}

\subsection{Experiments on Rain Streak Removal}
\input{rain.tex}

\subsection{Experiments on Image Dehazing}
\input{haze.tex}
\input{table_mixed}

\subsection{\maRevise{Cross-Task Evaluation}}

\maRevise{Due to similar problem formulation between different tasks of image restoration such as image reflection removal, image deraining and image dehazing, the methods designed for one task can be theoretically applied to other tasks. To compare the generalizability across different tasks of these ad hoc models for one specific task with the unified solution of our model, we perform cross-task evaluation in this set of experiments.
Specifically, we take one representative method with best performance for each task into comparison: \emph{ERRNet} for image reflection removal, \emph{PReNet} for image deraining and \emph{DM$^2$F-Net} for image dehazing. We evaluate their cross-task performance on three benchmarks, namely one benchmark for each of three tasks: Real20 for image reflection removal, Rain100H for image deraining and O-HAZE for image dehazing.  
}

\smallskip\noindent\maRevise{\textbf{Quantitative Evaluation.}}
\maRevise{Table~\ref{tab:mixed} presents the experimental results of cross-task evaluation. Our method consistently achieves the best performance among four methods across three tasks of image restoration, which manifests better generalizability of our model than other methods. All other methods (except our method) perform well on their specialized task but show poor performance on other two tasks. It is not surprising since our model is designed as a unified framework for different tasks of image restoration whilst other methods are designed for one specific task of image background restoration.
}

%% file: histogram.tex
\begin{table}[!t]
\centering
\caption{\maRevise{Ablation experiments on our \emph{DMGN} in iterms of \emph{PSNR} and \emph{SSIM} to investigate the effectiveness of each proposed technique in our \emph{DMGN} and the effect of each loss function.}}
\resizebox{0.9\linewidth}{!}{
\begin{tabular}{l| c c c c}
\toprule
\multirow{2}*{Methods}&\multicolumn{2}{c}{Real20~\cite{zhang2018single}} &\multicolumn{2}{c}{Wild~\cite{wan2017benchmarking}}   \\
\cmidrule(lr){2-3} 
\cmidrule(lr){4-5}

~ & PSNR & SSIM& PSNR & SSIM\\
\midrule
Base model&20.62&0.760&23.78&0.852\\
RDMC&21.96&0.792&24.42&0.882\\
R-RDMC&22.63&0.801&24.73&0.885\\
Coarse&22.49&0.798&24.12&0.867\\
\textbf{DMGN}&\textbf{23.05}&\textbf{0.823}&\textbf{25.18}&\textbf{0.894}\\
\midrule
 \maRevise{w/o} \maRevise{$\mathcal{L}_{\text{adv}}$}&\maRevise{22.66}&\maRevise{0.811}&\maRevise{24.05}&\maRevise{0.825}\\
\maRevise{w/o $\mathcal{L}_{p}$}&\maRevise{20.83}&\maRevise{0.747}&\maRevise{22.32}&\maRevise{0.796}\\
\textbf{Complete losses}&\textbf{23.05}&\textbf{0.823}&\textbf{25.18}&\textbf{0.894}\\
\bottomrule
\end{tabular}
}
\label{tab:ablation}
\end{table} 

%% file: table1.tex
\begin{table*}[!t]
\centering
\caption{Performance of different models for image reflection removal on four datasets in terms of \emph{PSNR} and \emph{SSIM}.}
\resizebox{0.75\linewidth}{!}{
\begin{tabular}{l| c c c c c c c c}
\toprule
\multirow{2}*{Methods}&\multicolumn{2}{|c}{Real20} &\multicolumn{2}{c}{Solid} &\multicolumn{2}{c}{Postcard}& \multicolumn{2}{c}{Wild}  \\
\cmidrule(lr){2-3} 
\cmidrule(lr){4-5}
\cmidrule(lr){6-7}
\cmidrule(lr){8-9}
~& PSNR & SSIM& PSNR & SSIM& PSNR & SSIM& PSNR & SSIM \\
\midrule
YW19~\cite{yang2019fast}&17.99&0.642&20.39&0.817&20.04&0.791&21.39&0.822\\
BDN~\cite{yang2018seeing}&18.97&0.740&22.73&0.853&20.71&0.856&22.34&0.821\\
PLNet~\cite{zhang2018single}&21.91&0.787&22.63&0.874&16.82&0.797&21.50&0.829\\
RmNet~\cite{wen2019single}&18.33&0.652&19.81&0.786&19.51&0.804&21.48&0.813\\
ERRNet~\cite{wei2019single}&22.86&0.798&24.69&0.891&21.85&0.878&25.03&0.874\\
\maRevise{IBCLN~\cite{li2019single}}&\maRevise{22.39}&\maRevise{0.798}&\maRevise{24.81}&\maRevise{0.890}&\maRevise{\textbf{23.38}}&\maRevise{0.875}&\maRevise{24.71}&\maRevise{0.883}\\
Zou \emph{et al.}~\cite{Zou_2020_CVPR}&22.75&0.791&23.21&0.882&18.98&0.742&\textbf{25.56}&0.892\\
\textbf{DMGN (ours)}&\textbf{23.05}&\textbf{0.823}&\textbf{25.13}&\textbf{0.904}&22.59&\textbf{0.902}&25.18&\textbf{0.894}\\
\bottomrule
\end{tabular}
}
\label{tab:comparison}
\end{table*} 

\begin{table}[!t]
\centering
\caption{\maRevise{Model complexity of our \emph{DMGN} and two state-of-the-art methods for image reflection removal in terms of GFLOPs and Inference time.}}
\renewcommand\arraystretch{1.2}
\resizebox{0.75\linewidth}{!}{
\begin{tabular}{l| c c}
\toprule
Model & GFLOPs & Inference time\\
\midrule
IBCLN~\cite{li2019single}&289.2&0.071s\\
ERRNet~\cite{wei2019single}&371.9&0.074s\\
\midrule
DMGN (ours)&232.5&0.066s\\
\bottomrule
\end{tabular}
}
\label{tab:efficiency}
\end{table}

%% file: human_table.tex
\begin{table}[!t]
\centering
\caption{Human evaluation on the reflection removal results. 50 human subjects are asked to perform comparison between our model and ERRNet on the restored background images of 50 randomly test samples. Our model obtains $53.2\%$ among $50\times 50 = 2500$ votes and wins on 42 samples compared to other 3 state-of-the-art methods.}
\renewcommand\arraystretch{1.4}
\resizebox{0.9\linewidth}{!}{
\begin{tabular}{l| c c}
\toprule
~ & Share of the vote & Winning samples\\
\midrule
ERRNet~\cite{wei2019single}&\maRevise{25.6\%}&\maRevise{2}\\
\maRevise{IBCLN~\cite{li2019single}} & \maRevise{19.0\%}& \maRevise{4}\\
\maRevise{Zou et al.~\cite{Zou_2020_CVPR}} & \maRevise{2.2\%}& \maRevise{0}\\
DMGN (ours)&\maRevise{53.2\%}&\maRevise{42}\\
\bottomrule
\end{tabular}
}
\label{tab:human}
\end{table} 

%% file: rain.tex
Next we conduct experiments to evaluate the performance of our \emph{DMGN} on rain streak removal.

\smallskip\noindent\textbf{Datasets.} We perform experiments on two popular benchmark datasets for rain streak removal: Rain100H~\cite{yang2017deep} and Rain100L~\cite{yang2017deep}. Rain100H is composed of heavy rainy images, among which 1800 and 100 images are split for training and test respectively. By contrast, Rain100L is collected for light rainy cases and contains 200 images for training and 100 images for test. Similar to experiments of 
image reflection removal, we also adopt \emph{PSNR} and \emph{SSIM} as evaluation metrics here.
Further, we collect a small set of real-world rain streak images for qualitative evaluation.

\smallskip\noindent\textbf{Baselines}. We compare our \emph{DMGN} model with state-of-the-art methods \miRevise{for image rain streak removal} including: \textbf{DDN}~\cite{fu2017removing} which focuses on preserving texture details, \textbf{JORDER}~\cite{yang2017deep} that aims to perform rain streak detection and removal jointly by a multi-stream network, \textbf{RGN}~\cite{fan2018residual} which designs a lightweight cascaded network, \textbf{RESCAN}~\cite{li2018recurrent} which seeks to model the rain streak layer precisely to obtain the background image in a round-about way, and \textbf{PReNet}~\cite{ren2019progressive} synthesizing the background image by a recurrent structure. \miRevise{Besides, we also evaluate the performance of rain streak removal by \textbf{AttentiveGAN}~\cite{qian2018attentive}, the state-of-the-art method specifically designed for raindrop removal which learns an attention map to locate the raindrops and guide the raindrop removal.}

\begin{figure*}[!tp]
  \centering 
  \begin{minipage}[b]{0.95\linewidth}
  \includegraphics[width=\linewidth]{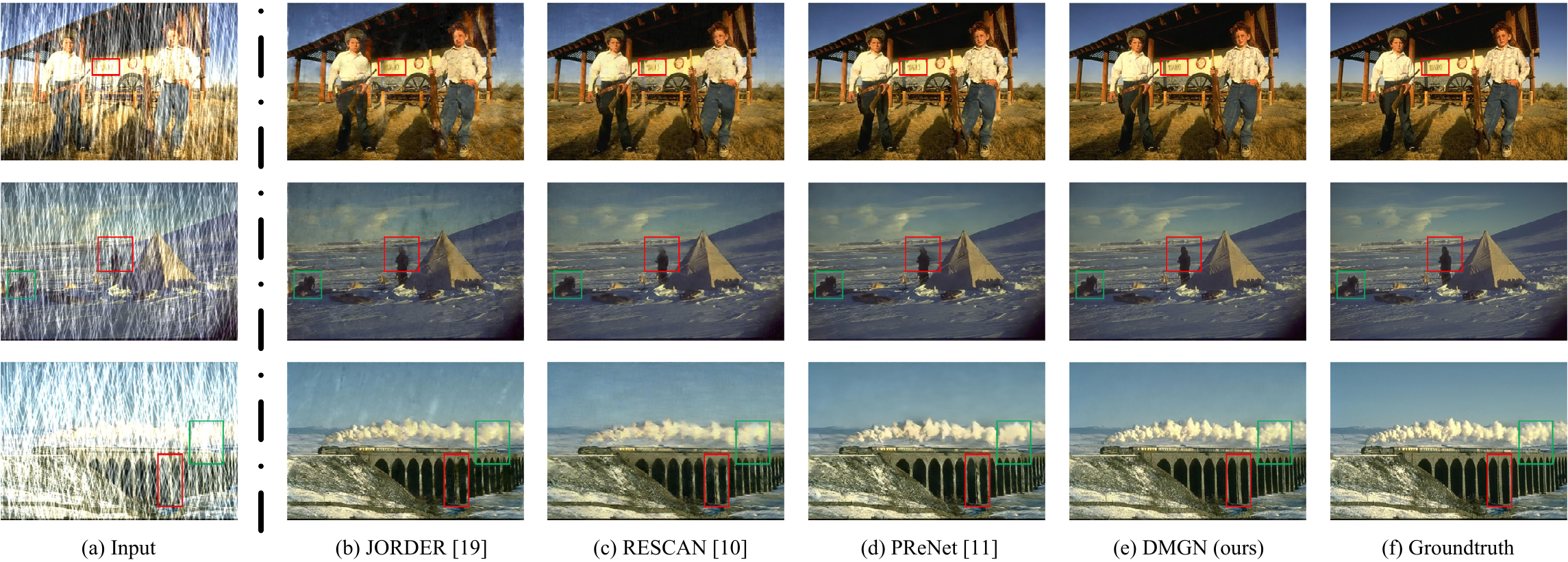}\vspace{4pt}
  \end{minipage}
    \vspace{-10pt}
\caption{Qualitative comparison of rain streak removal between different methods
on three randomly selected examples from test set of Rain100H [32] dataset. Our model is able to restore cleaner and higher-quality background image than other methods, particularly in the regions indicated by bounding boxes. Best viewed in zoom-in mode.}
  \vspace{-8pt}
\label{fig:rain-synthetic}
\end{figure*}

\begin{figure*}[!tp]
  \centering 
  \begin{minipage}[b]{0.75\linewidth}
  \includegraphics[width=\linewidth]{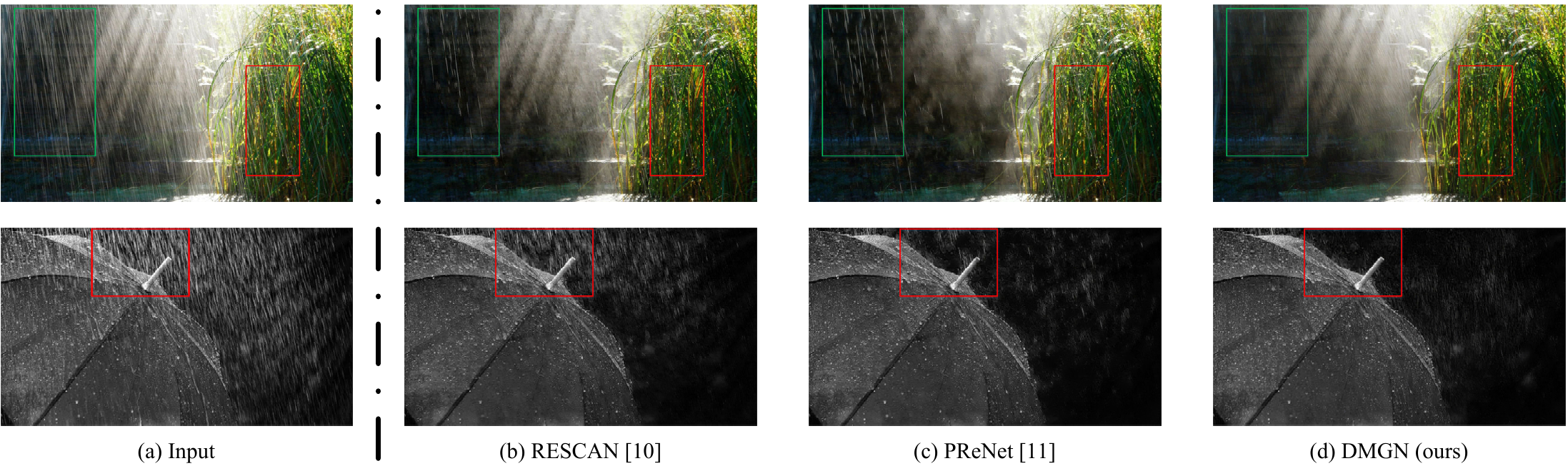}\vspace{4pt}
  \end{minipage}
    \vspace{-4pt}
\caption{Qualitative comparison of rain streak removal between different methods on two real-world images. Our \emph{DMGN} performs best among all methods and is able to restore much cleaner background in both cases than other methods. Best viewed in zoom-in mode.}
  \vspace{-4pt}
\label{fig:rain-realworld}
\end{figure*}

\input{table2.tex}

\begin{figure*}[!t]
  \centering 
  \begin{minipage}[b]{0.93\linewidth}
  \includegraphics[width=\linewidth]{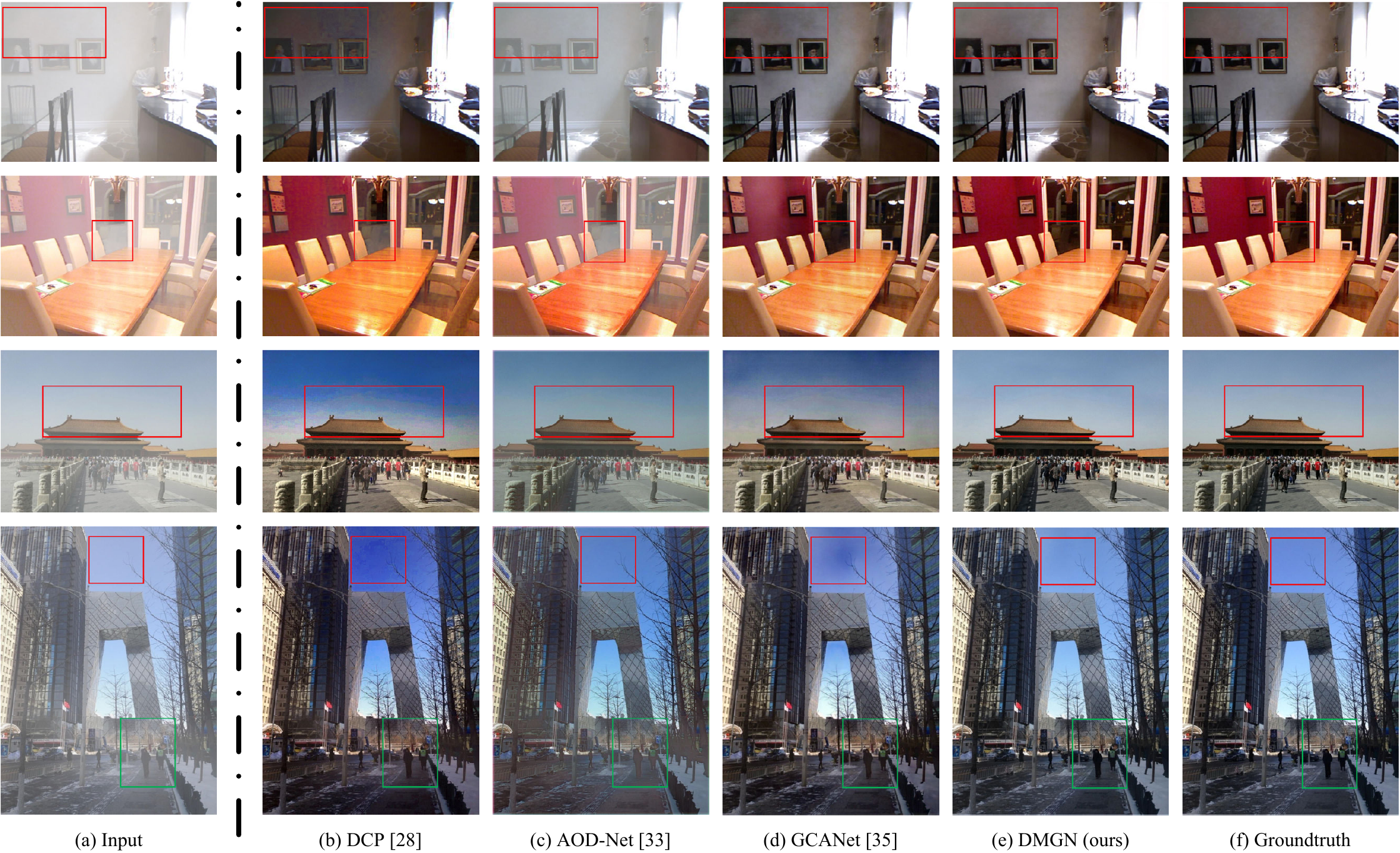}\vspace{4pt}
  \end{minipage}
    \vspace{-4pt}
\caption{Visualization of restored background images by five models for image dehazing on three randomly selected samples from test set. Our model is able to restore cleaner and higher-quality background image than other methods, particularly in the regions indicated by bounding boxes. Best viewed in zoom-in mode.}
  \vspace{-4pt}
\label{fig:dehazing}
\end{figure*}

\smallskip\noindent\textbf{Quantitative Evaluation.} Table~\ref{tab:rain_comparison} presents the quantitative results of different models for rain streak removal. It manifests that our model achieves the best results \miRevise{and surpasses other models by a large margin} on both Rain100H (heavy-rain cases) and Rain100L (light-rain cases). In particular, our \emph{DMGN} performs distinctly better than two most powerful methods for rain streak removal: \emph{RESCAN}~\cite{li2018recurrent} which focuses on modeling the rain streak layer and \emph{PReNet}~\cite{ren2019progressive} employing the same customized module recurrently for background restoration.

It is worth noting that the \emph{AttentiveGAN}~\cite{qian2018attentive}, which is a state-of-the-art method for raindrop removal, performs fairly inferior to our \emph{DMGN} on rain streak removal. We observe that the training of its attention map is highly prone to collapse since the supervised training of the attention map is very sensitive to the threshold value for ground truth of rain mask, which is extremely hard to tune in heavy rain cases (Rain100H). It reveals the difficulties of generalization across different noise patterns for \emph{AttentiveGAN}.

\smallskip\noindent\textbf{Qualitative Evaluation.} To have a qualitative evaluation, we visualize the restored background images by different models on two randomly selected examples from Rain100H dataset in Figure~\ref{fig:rain-synthetic}. It is manifested that the rain streaks are removed more thoroughly by our \emph{DMGN} than other models, especially in the regions indicated by bounding boxes.

\smallskip\noindent\textbf{Evaluation on real-world samples.} As a complementary experiment to experiments on synthetic data, we also present two real-world examples randomly selected to qualitatively show the comparison between our model and two most powerful models for rain streak removal. For a fair comparison, all models are pre-trained on Rain100H dataset. Figure~\ref{fig:rain-realworld} shows that our model performs better than other methods. Note that there are several visible rain streaks that \emph{RESCAN} and \emph{PReNet} cannot remove whereas our model successfully removes them.

%% file: table2.tex
\begin{table}[!t]
\centering
\caption{Performance of different models for image rain streak removal on two datasets in terms of \emph{PSNR} and \emph{SSIM}.}
\renewcommand\arraystretch{1.2}
\resizebox{0.9\linewidth}{!}{
\begin{tabular}{l| c c c c}
\toprule
\multirow{2}*{Methods}&\multicolumn{2}{c}{Rain100L} &\multicolumn{2}{c}{Rain100H}   \\
\cmidrule(lr){2-3} 
\cmidrule(lr){4-5}

~ & PSNR & SSIM& PSNR & SSIM\\
\midrule
DDN~\cite{fu2017removing}&32.16&0.936&21.92&0.764\\
RGN~\cite{fan2018residual}&33.16&0.963&25.25&0.841\\
JORDER~\cite{yang2017deep}&36.61&0.974&26.54&0.835\\
DID~\cite{zhang2018density}&36.14&0.941&26.12&0.833\\
AttentiveGAN~\cite{qian2018attentive}&35.36&0.977&28.50&0.883\\
RESCAN~\cite{li2018recurrent}&37.14&0.977&28.88&0.866\\
PReNet~\cite{ren2019progressive}&37.48&0.979&29.46&0.899\\
\textbf{DMGN (ours)}&\textbf{38.53}&\textbf{0.983}&\textbf{30.72}&\textbf{0.903}\\
\bottomrule
\end{tabular}
}
\vspace{-15pt}
\label{tab:rain_comparison}
\end{table}

%% file: haze.tex
In this section we conduct experiments to evaluate the performance of our \emph{DMGN} on image dehazing.

\smallskip\noindent\textbf{Datasets.} We perform experiments on three standard benchmark datasets for image dehazing, including RESIDE~\cite{li2018benchmarking}, O-HAZE~\cite{O-HAZE} and TestA-DCPDN~\cite{zhang2018densely}. 

\input{table3.tex}

RESIDE is a synthetic dataset derived from NYU V2~\cite{silberman2012indoor} and Middlebury Stereo dataset~\cite{scharstein2003high}. It consists of two types of scenes: indoor and outdoor. The indoor training set includes 13990 hazy images which are obtained from 1399 clean indoor images by adding haze based on the atmosphere light atmosphere scattering model~\cite{narasimhan2002vision}. The outdoor training set contains 72135 hazy images which are generated from 2061 clean outdoor images. Following the routine setting~\cite{li2018benchmarking}, we use the Synthetic Objective Testing Set (SOTS) in RESIDE for evaluation. This test dataset contains 500 indoor images and 500 outdoor images. Similar to former experiments, we also adopt \emph{PSNR} and \emph{SSIM} as evaluation metrics \miRevise{for image dehazing}.

\maRevise{O-HAZE is a outdoor-scene dataset consisting of pairs of real hazy and corresponding haze-free images. The real hazy images were captured by professional haze machines in presence of real haze. O-HAZE contains images 45 different outdoor scenes depicting the same visual content recorded in both haze-free and hazy conditions respectively, under the same illumination parameters.}

\maRevise{TestA-DCPDN is a synthetic dataset which is also collected from NYU dataset~\cite{silberman2012indoor} by randomly setting different atmospheric light conditions
and the scattering coefficient to synthesize haze.
The training set comprises of 4000 images while the test set (named TestA) contains 400 images.}

\smallskip\noindent\textbf{Baselines.} We compare our model with following state-of-the-art methods for image dehazing:
\textbf{DCP}~\cite{he2010single} which relies on the assumed pixel statistic prior of color channels; \textbf{DehazeNet}~\cite{cai2016dehazenet} that is an end-to-end model to predict medium transmission map and \miRevise{restore} clean background image through the atmosphere scattering model~\cite{narasimhan2002vision}; \textbf{AOD-Net}~\cite{li2017aod} which designs a specific module to estimate physical model's parameters; \textbf{GFN}~\cite{ren2018gated} which employs multi-scale refining scheme to estimate three haze residual maps and then generate final haze-free image; \textbf{GCANet}~\cite{chen2019gated} which proposes a smoothed dilated operation and gated fusion structure to improve restoration quality;
\maRevise{\textbf{DA}~\cite{shao2020domain} which addresses the image dehazing in a domain adaptation paradigm to bridge the gap between synthetic and real domains;}
\maRevise{\textbf{DM$^2$F-Net}~\cite{deng2019deep} explores the potentials of multi-layer features of deep CNNs by designing a specific attention mechanism;}
\maRevise{\textbf{DCPDN}~\cite{zhang2018densely} which aims to jointly learn the transmission map, atmospheric light and dehazing all together in a physics-driven manner.}

\smallskip\noindent\textbf{Quantitative Evaluation.} Table~\ref{tab:dehazing} presents the quantitative evaluation results of different models for image dehazing. Our \emph{DMGN} outperforms all other state-of-the-art models by a large margin in all three benchmarks, which demonstrates the distinct advantages of our model over other models.

\smallskip\noindent\textbf{Qualitative Evaluation.} To have a qualitative evaluation, we visualize the restored background images by different models for image dehazing on three randomly selected examples from `SOTS' test set in indoor and outdoor scenes in Figure~\ref{fig:dehazing}. It shows that our \emph{DMGN} is able to remove the haze more thoroughly by our \emph{DMGN} than other methods. Although the model `DCP' restores a relatively clean background, it alters the color distribution of background image which is far from the groundtruth image, \miRevise{presumably} due to its assumption of dark channel prior.

%% file: table3.tex
\begin{table*}[!t]
\centering
\caption{Performance of different models for image dehazing on three datasets in terms of \emph{PSNR} and \emph{SSIM}.}
\renewcommand\arraystretch{1.2}
\begin{tabular}{l| c c c c c c c c}
\toprule
\multirow{2}*{Methods}&\multicolumn{2}{c}{Indoor (SOTS)} &\multicolumn{2}{c}{Outdoor (SOTS)}&\multicolumn{2}{c}{\maRevise{O-HAZE}} &\multicolumn{2}{c}{\maRevise{TestA-DCPDN}}   \\
\cmidrule(lr){2-3} 
\cmidrule(lr){4-5}
\cmidrule(lr){6-7}
\cmidrule(lr){8-9}

~ & PSNR & SSIM& PSNR & SSIM& PSNR & SSIM& PSNR & SSIM\\
\midrule
DCP~\cite{he2010single}&20.44&0.864&20.84&0.862 & \maRevise{16.78}&\maRevise{0.635}&\maRevise{16.62}&\maRevise{0.818}\\
DehazeNet~\cite{cai2016dehazenet}&21.34&0.875&22.60&0.861&\maRevise{16.24}&\maRevise{0.669}&\maRevise{21.14}&\maRevise{0.850}\\
AOD-Net~\cite{li2017aod}&19.06&0.851&20.29&0.877&\maRevise{15.03}&\maRevise{0.539}&\maRevise{20.86}&\maRevise{0.879}\\
GFN~\cite{ren2018gated}&22.30&0.880&21.55&0.845&\maRevise{18.16}&\maRevise{0.671}&\maRevise{22.30}&\maRevise{0.880}\\
GCANet~\cite{chen2019gated}&30.23&0.980&22.80&0.920&\maRevise{16.28}&\maRevise{0.645}&\maRevise{27.23}&\maRevise{0.921}\\
\maRevise{DA~\cite{shao2020domain}}&\maRevise{25.45}&\maRevise{0.918} &\maRevise{26.07} & \maRevise{0.905}&\maRevise{19.89} & \maRevise{0.812}&\maRevise{25.87}&\maRevise{0.924}\\
\maRevise{DM$^2$F-Net~\cite{deng2019deep}} &\maRevise{34.35}& \maRevise{0.972}& \maRevise{25.46}& \maRevise{0.923}& \maRevise{23.46}& \maRevise{0.851}& \maRevise{35.67}& \maRevise{0.978}\\
\maRevise{DCPDN~\cite{zhang2018densely}} & \maRevise{15.97}& \maRevise{0.823}& \maRevise{20.01}& \maRevise{0.849}& \maRevise{22.80}& \maRevise{0.744}& \maRevise{29.29}& \maRevise{0.946}\\
\textbf{DMGN (ours)}&\textbf{35.33}&\textbf{0.986}&\textbf{28.99}&\textbf{0.971}&\maRevise{\textbf{25.11}}&\maRevise{\textbf{0.910}}&\maRevise{\textbf{36.35}}&\maRevise{\textbf{0.987}}\\
\bottomrule
\end{tabular}
\label{tab:dehazing}
\end{table*}

%% file: table_mixed.tex
\begin{table}[!t]
\centering
\caption{\maRevise{Cross-task evaluation across three tasks of image background restoration on different methods. One representative method for each of three tasks is selected to report their performance on three datasets in terms of \emph{PSNR} and \emph{SSIM}. Real20, Rain100H and O-HAZE are benchmarks for image reflection removal, image rain streak removal and image dehazing respectively.}}
\renewcommand\arraystretch{1.2}
\begin{tabular}{l| c c c c c c}
\toprule
\multirow{3}*{Methods}&\multicolumn{2}{c}{\maRevise{Real20}} &\multicolumn{2}{c}{\maRevise{Rain100H}}&\multicolumn{2}{c}{\maRevise{O-HAZE}}\\
\cmidrule(lr){2-3} 
\cmidrule(lr){4-5}
\cmidrule(lr){6-7}

~ & \maRevise{PSNR} & \maRevise{SSIM}& \maRevise{PSNR} & \maRevise{SSIM}& \maRevise{PSNR} & \maRevise{SSIM}\\
\midrule
\maRevise{ERRNet~\cite{wei2019single}}&\maRevise{22.86}&\maRevise{0.798}&\maRevise{22.78}&\maRevise{0.802} & \maRevise{21.60}&\maRevise{0.795}\\
\maRevise{PReNet~\cite{ren2019progressive}}&\maRevise{19.70}&\maRevise{0.742}&\maRevise{29.46}&\maRevise{0.899} & \maRevise{20.47}&\maRevise{0.777}\\
\maRevise{DM$^2$F-Net~\cite{deng2019deep}}&\maRevise{20.01}&\maRevise{0.761}&\maRevise{21.97}&\maRevise{0.749} & \maRevise{23.46}&\maRevise{0.851}\\
\maRevise{\textbf{DMGN (ours)}} &\maRevise{\textbf{23.05}}&\maRevise{\textbf{0.823}}&\maRevise{\textbf{30.72}}&\maRevise{\textbf{0.903}}&\maRevise{\textbf{25.11}}&\maRevise{\textbf{0.910}}\\
\bottomrule
\end{tabular}
\label{tab:mixed}
\end{table} 

%% file: conclusion.tex
In this work we have presented Deep-Masking Generative Network (\emph{DMGN}), a unified framework for background restoration from superimposed images. Our \emph{DMGN} iteratively employs the Residual Deep-Masking Cell, which is designed to control the information propagation, to generate the image in a progressive-refining manner. We further propose a two-pronged strategy to effectively leverage the generated noise image as contrasting cues for refining the background image. Extensive validation on three tasks demonstrates the well generalization across different noise patterns. In particular, our \emph{DMGN} outperforms other ad hoc state-of-the-art methods specifically designed for each single task.

\maRevise{Performing image restoration for nighttime images remains a challenging task for our model as well as other existing methods for image restoration. This is primarily because smaller light contrast in night time increases the difficulty of separating the background from the noise, which will be studied in our future work.}